\documentclass{article}

\usepackage{hyperref} 
\hypersetup{
colorlinks=true, 
breaklinks=true,
urlcolor= blue, 
linkcolor= blue,
citecolor=red, 
}

\usepackage{multicol}

\usepackage{color}

\usepackage{xspace}
\newcommand{\matpower}[0]{\textsc{Matpower}\xspace}

\usepackage{url}
\usepackage{amssymb,amsmath, amsthm,bm}
\usepackage{graphicx}
\usepackage{authblk}

\begin{document}

\vspace*{\fill}
\begin{center}
{\Large \bf Note to the Reader}\\
\vspace{0.5cm}
\begin{minipage}{.6\textwidth}
The NESTA archive has been discontinued in favor of the official IEEE PES OPF Benchmark Library, PGLib-OPF (\href{https://arxiv.org/abs/1908.02788}{report}, \href{https://github.com/power-grid-lib}{data}).  Future works are encouraged to use the PGLib-OPF archive.  Access to the NESTA archive is provided in the interest of replicating previous works.  For access please contact the administrator: \\
Carleton Coffrin, carleton@coffrin.com \\
\vspace{6cm}
\end{minipage}
\end{center}
\vfill 

\clearpage
\title{NESTA \\ The {\sc Nicta} Energy System Test Case Archive}  
\author[1,2,3]{Carleton Coffrin} 
\author[1]{Dan Gordon} 
\author[1,2]{Paul Scott}
\affil[1]{Optimisation Research Group, NICTA}
\affil[2]{College of Engineering and Computer Science, Australian National University}
\affil[3]{Computing and Information Systems, University of Melbourne}
\maketitle

\abstract{
In recent years the power systems research community has seen an explosion of work applying operations research techniques to challenging power network optimization problems.  Regardless of the application under consideration, all of these works rely on power system test cases for evaluation and validation.  However, many of the well established power system test cases were developed as far back as the 1960s with the aim of testing AC power flow algorithms.  It is unclear if these power flow test cases are suitable for power system optimization studies.  This report surveys all of the publicly available AC transmission system test cases, to the best of our knowledge, and assess their suitability for optimization tasks.  It finds that many of the traditional test cases are missing key network operation constraints, such as line thermal limits and generator capability curves.  To incorporate these missing constraints, data driven models are developed from a variety of publicly available data sources.  The resulting extended test cases form a compressive archive, NESTA, for the evaluation and validation of power system optimization algorithms.
}

\section*{Nomenclature}
\begin{multicols}{2} 

\begin{description}
  \item [{$V = v \angle \theta$}]  - AC voltage
  \item [{$p+iq$}] - AC power
  \item [{$e$}] - AC current magnitude
  \item [{$y$}] - Line admittance magnitude
  \item [{$r+ix$}] - Line impedance
  \item [{$b^c$}] - Line charge
  \item [{$t$}] - Line apparent power thermal limit
  \item [{$p^d+iq^d$}] - AC power demand
  \item [{$p^g+iq^g$}] - AC power generation
  \item [{$c_0,c_1,c_2$}] - Generation cost coefficients 
  \item [{$\dot{v}$}] - Nominal voltage
  \item [{$\theta^\Delta$}] - Phase angle difference
  \item [{$f$}] - Fuel category
  \item [{$\mu$}] - Mean
  \item [{$\sigma$}] - Standard deviation
  \item [{$\lambda$}] - Rate of an exponential distribution
    \item [{$\bm e$}] - Exponential constant
  \item [{$x^u$}] - Upper bound of $x$
  \item [{$x^l$}] - Lower bound of $x$
  \item [{$\hat{x}$}] - An estimation of $x$
  \item [{$\boldsymbol x$}] - A constant value
\end{description}

\end{multicols}

\section{Introduction}

Over the last decade, power systems research has experienced an explosion of publications applying operations research technologies to power network optimization problems, such as economic dispatch \cite{Rafael_analysisand}, optimal power flow \cite{744492,744495}, optimal transmission switching \cite{4492805}, and network expansion planning \cite{6407493}, just to name a few.  Many of these publications have been inspired by new formulations of the AC power flow equations, such as the SDP/QC/SOCP relaxations \cite{Bai2008383,QCarchive,Jabr06} and the DC/LPAC/IV-Flow approximations \cite{Stott_2009bb,LPAC_ijoc,IVModel}, which bring industrial-strength general-purpose optimization techniques to AC power system applications.  


Independent of the specific application or the power flow model under consideration, all of these works require AC power network data for experimental validation and it is important that realistic data is used to validate these applications and solution techniques.  However, due to the sensitive nature of critical infrastructure, this kind of data is often very difficult to obtain.  As a result, the research community most often uses a collection of power networks that are distributed with the \matpower research platform \cite{matpower}.  Many of these network models are over thirty years old \cite{IEEEBench}. They were originally designed to test the AC power flow problem, which is a feasibility task rather than an optimization task.  Consequently, these test cases are missing key pieces of information, such as line thermal limits and generator cost functions, which are critical in optimization applications. Incorporating this missing information requires careful consideration, as it is easy to make test cases that are either trivial or infeasible.

This report has three goals: Firstly to conduct an extensive survey of all publicly available transmission system data sets and to curate them in one place with a consistent data format.  Secondly, to make all of these test cases usable for optimization studies by reconstructing the missing data, and thirdly, to demonstrate that these completed test cases provide a valuable building block for developing new test cases tailored to specific application domains.  The result is a comprehensive transmission network test case archive designed to be used for AC power flow optimization studies.  

We begin, in Section \ref{sec:motivation}, with a discussion of why the \matpower cases are inadequate for the study of optimization applications. We next conduct, in Section \ref{sec:networks}, a boarder survey of publicly available network data.  Recognizing that the majority of test cases are missing key pieces of information, Section \ref{sec:data_models} proposes data driven approaches for reconstructing this information for the available test cases.  Finally, in Section \ref{sec:nesta} we develop new test cases, and demonstrate that they are indeed interesting for testing optimization applications. Section \ref{sec:conclusion} concludes the report.


\section{Motivation}
\label{sec:motivation}

To better understand the \matpower test cases and keep this report within a reasonable scope, we focus on the seminal power network optimization problem: AC Optimal Power Flow (AC-OPF) \cite{744492,744495}.  The AC-OPF problem is a  continuous non-convex optimization problem and like many AC optimization problems, it can be shown to be NP-hard \cite{Verma09, 5971792, ACSTAR2015}.  To address the computational difficulty of the AC-OPF problem, convex relaxations have been proposed \cite{Bai2008383,QCarchive,Jabr06,6102366} as computationally efficient alternatives to the AC power flow equations.

For any given AC-OPF test case, we can measure the difficulty of the problem by comparing a heuristic solution of the true problem to lower bounds obtained by convex relaxations.  Challenging test cases exhibit a large {\em optimality gap}, that is the relative difference between the objective value of the heuristic and relaxation as in \eqref{eq:opt_gap}.
\begin{align}
\frac{\text{AC Heuristic} - \text{Relaxation}}{\text{AC Heuristic}} \label{eq:opt_gap}
\end{align}

\begin{table}[t!]
\center
\begin{tabular}{|r||r||r|r|r|r|r|r|r|r|r||r|r|r|r|c|c|}
\hline
                   & \$/h & \multicolumn{4}{c|}{Optimality Gap (\%)}  \\
Test Case & AC  & CP & NF+LL & SOC & SDP \\
\hline
\hline
 case4gs & 17318.30 & 0.99 & 0.05 & 0.00 & \bf 0.00  \\
\hline
 case5 & 17551.89 & 15.62 & 14.55 & 14.54 & 5.22 \\
\hline
 case6bus & 3844.12 & 0.33 & 0.01 & 0.00 & \bf 0.00  \\
\hline
 case6ww & 3143.97 & 3.10 & 0.81 & 0.63 & \bf 0.00  \\
\hline
 case9 & 5296.69 & 1.52 & 0.17 & 0.00 & 0.00 \\
\hline
 case14 & 8081.53 & 5.43 & 0.20 & 0.08 & \bf 0.00  \\
\hline
 case24\_ieee\_rts & 63352.20 & 3.71 & 0.36 & 0.01 & \bf 0.00  \\
\hline
 case30 & 576.89 & 2.03 & 0.69 & 0.57 & 0.00 \\
\hline
 case\_ieee30 & 8906.14 & 6.32 & 0.24 & 0.04 & \bf 0.00  \\
\hline
 case39 & 41864.18 & 1.43 & 0.08 & 0.02 & 0.00 \\
\hline
 case57 & 41737.79 & 1.75 & 0.25 & 0.06 & \bf 0.00  \\
\hline
 case89pegase & 5819.81 & 1.50 & 0.36 & 0.17 & err. \\
\hline
 case118 & 129660.69 & 2.86 & 0.35 & 0.25 & 0.00 \\
\hline
 case300 & 719725.08 & --- & 0.34 & 0.15 & 0.00 \\
\hline
 case1354pegase & err. & n/a & n/a & n/a & n/a \\
\hline
 case2383wp & 1868511.78 & 5.35 & 1.38 & 1.05 & 0.37 \\
\hline
 case2736sp & 1307883.11 & 2.44 & 0.46 & 0.30 & 0.00 \\
\hline
 case2737sop & 777629.29 & 1.75 & 0.39 & 0.25 & 0.00 \\
\hline
 case2746wp & 1631775.07 & 3.09 & 0.53 & 0.32 & 0.00 \\
\hline
 case2746wop & 1208279.78 & 2.50 & 0.56 & 0.37 & 0.00 \\
\hline
 case2869pegase & 133999.84 & 1.16 & 0.19 & 0.09 & err. \\
\hline
 case3012wp & 2591706.52 & --- & 4.92 & 0.78 & err. \\
\hline
 case3120sp & 2142703.73 & --- & 5.20 & 0.53 & err. \\
\hline
 case3375wp & 7412030.61 & --- & 1.74 & 0.26 & err. \\
\hline
 case9241pegase & err. & n/a & n/a & n/a & n/a \\
\hline
\end{tabular}
\caption{AC-OPF Bounds on the \matpower Test Cases.}
\label{tbl:ac:bounds}
\end{table}

\noindent
To investigate the difficulty of the AC-OPF problem on the \matpower test cases, we compare, in Table \ref{tbl:ac:bounds}, the following five variants of the AC-OPF problem,  
\begin{enumerate}
\item AC -- a heuristic solution to the full non-convex problem
\item CP --  a {\em copper plate} relaxation which ignores all of the network constraints 
\item NF+LL --  a network flow relaxation incorporating a convexification of line losses \cite{pscc_ots}
\item SOC -- the second-order cone relaxation \cite{Jabr06}
\item SDP -- the semi-definite programming relaxation \cite{Bai2008383}
\end{enumerate}
%
The first four results were computed using IPOPT 3.12 \cite{Ipopt} and the fifth result comes from a state-of-the-art SDP-OPF implementation \cite{opfBranchDecompImpl} using the SDPT3 4.0 \cite{Toh99sdpt3} solver.  
Note that the copper plate model cannot be applied to a few networks because they feature negative $r$ and $x$ values.

The results for the copper plate model indicate that the network flow constraints typically only increase the AC-OPF objective by 2\%--5\% in these test cases.  If line losses are incorporated (i.e. NF+LL and SOC), then the optimality gap can be reduced further, to less than 1\% in most cases.  Applying the more computationally expensive SDP relaxation indicates that these heuristic solutions are near global optimality for all but one of the test cases, as first observed in \cite{5971792}.  It is also interesting to note that \cite{pscc_ots} showed NF+LL is also a relaxation of the AC Optimal Transmission Switching problem (AC-OTS), indicating that these test cases are trivial for that optimization problem as well.  

\subsection{A 3-Bus AC-OPF Case Study}
\label{sec:case_study}

The small optimality gaps observed in the previous section suggests that the non-convex nature of the AC-OPF problem is not a major issue in the \matpower test cases.  To illustrate that small optimality gaps are not a fundamental property of the AC-OPF problem, this section develops a simple 3-bus test case with significant optimality gaps.  It also illuminates how the optimality gaps are sensitive to the input data, specifically line thermal limits and generator cost functions.  

\begin{figure}[t]
\center
    \includegraphics[width=4.0cm]{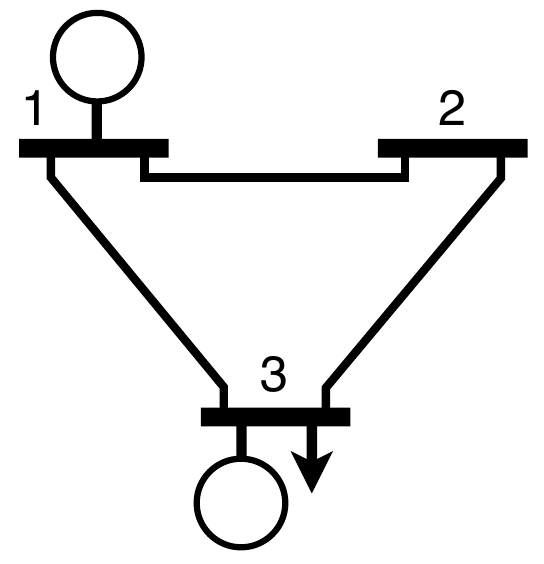} 
\caption{3-Bus Example Network Diagram.}
\label{fig:3_bus_example}
\end{figure}

The three-bus power system is depicted in Figure \ref{fig:3_bus_example} and the associated network parameters are presented in Table \ref{tbl:3_bus_network_data}.  This network is designed with a distant cheap generator at bus 1 and an expensive generator at bus 3, which is co-located with the load.  The cheapest solution delivers as much power as possible from generator 1 and as little as possible from generator 3.  Based on the network demand of 100 MW, we can observe that the cost of any AC-OPF solution is within the range of \$100--\$1,000.
The AC-OPF solution is presented in Table \ref{tbl:3_bus_network_solution_closed}.
Without any binding constraints, all of the load is served from the distant generator with a cost of \$101, with the small increase above \$100 being due to line losses.

\begin{table}[t!]
\centering
\begin{tabular}{|c||c|c|c|c|}
\hline
\multicolumn{5}{|c|}{Bus Parameters} \\
\hline
Bus & $\boldsymbol {p^d}$ & $\boldsymbol {q^d}$ & $\boldsymbol {v^l}$ & $\boldsymbol {v^u}$ \\
\hline
1 & 0 & 0 & 0.9 & 1.1 \\ 
\hline 
2 & 0 & 0 & 0.9 & 1.1 \\ 
\hline 
3 & 100 & 0 & 0.9 & 1.1 \\ 
\hline 
\end{tabular}
\vspace{0.5cm}
\begin{tabular}{|c||c|c|c|c|}
\hline
\multicolumn{5}{|c|}{Line Parameters} \\
\hline
From--To bus & $\boldsymbol {r}$ & $\boldsymbol {x}$ & $\boldsymbol {b^c}$ & $\boldsymbol {t}$ \\
\hline
1--2 & 0.00 & 0.05 & 0 & $\infty$ \\ 
\hline 
2--3 & 0.00 & 0.05 & 0 & $\infty$ \\ 
\hline 
1--3 & 0.10 & 0.10 & 0 & $\infty$ \\ 
\hline 
\end{tabular}
\begin{tabular}{|c||c|c||c|c|c|}
\hline
\multicolumn{6}{|c|}{Generator Parameters} \\
\hline
Generator & $\boldsymbol {p^{gl}}, \boldsymbol {p^{gu}}$ & $\boldsymbol {q^{gl}}, \boldsymbol {q^{gu}}$ & $\boldsymbol {c_2}$ & $\boldsymbol {c_1}$ & $\boldsymbol {c_0}$ \\
\hline
1 & $0,\infty$ & $-\infty,\infty$ & 0 & 1 & 0 \\ 
\hline 
3 & $0,\infty$ & $0,0$ & 0 & 10 & 0 \\ 
\hline 
\end{tabular}

\caption{Three-Bus System Network Data (100 MVA Base).}

\label{tbl:3_bus_network_data}
\end{table}

\begin{table}[t!]
\centering
\begin{tabular}{|c||c|c|c|c|}
\hline
\multicolumn{5}{|c|}{Bus Values} \\
\hline
Bus & $p^g$ & $q^g$ & $v$ & $\theta$, deg. \\
\hline
1 & 101 & 5 & 1.100 & 0.000 \\ 
\hline 
2 & 0 & 0 & 1.090 & -1.434 \\ 
\hline 
3 & 0 & 0 & 1.080 & -2.895 \\ 
\hline 
\end{tabular}
\begin{tabular}{|c||c|c|}
\hline
\multicolumn{3}{|c|}{Line Values} \\
\hline
Line & From MVA & To MVA\\
\hline
1--2 & 64 & 63 \\ 
\hline 
2--3 & 63 & 63 \\ 
\hline 
1--3 & 44 & 44 \\ 
\hline 
\end{tabular}

\caption{Three-Bus System Solution.}
\label{tbl:3_bus_network_solution_closed}
\end{table}

 The solution gets more interesting when binding network constraints (i.e. congestion) are introduced. We consider two congestion scenarios for this 3-bus example: (1)
line capacity congestion, introduced by reducing the thermal limit of line 2--3 to
1 MVA and (2) voltage bound congestion, introduced by reducing the voltage range from
$\pm 0.1$ to $\pm 0.02$. Table \ref{tbl:3_bus_results} reports the
cheapest generator dispatch across the different scenarios and power flow models. 
The table illustrates that the various power flow relaxations (CP, NF+LL, SOC, SDP) form 
a gradient of different accuracies, which are positively correlated with their computational difficulty. It should also be noted that by manipulating the generator costs (i.e. $\boldsymbol c_{1,1}$, $\boldsymbol c_{1,3}$), the optimality gap may be increased arbitrarily.

This 3-bus example is contrived, and such extreme cases are unlikely to occur in practice.  One may wonder if the \matpower test cases exhibit small optimally gaps simply because they are more realistic.  Upon a detailed investigation, we find that the small optimality gaps can be attributed to the following features of the networks:
\begin{enumerate}
\item Many of the line thermal limits are set to large non-binding values (i.e. 9900 MVA).
\item The synchronous condensers in some test cases were converted into active power generators.
\item Generators are assigned quadratic cost curves, often with the same coefficients.
\end{enumerate}
Throughout this report, we strive to produce test cases that are consistent with real-world data sources whenever possible.  Interestingly, the resulting test cases strike a balance between the \matpower test cases and this extreme 3-bus example. 


\begin{table}[t!]
\centering
\begin{tabular}{|r||r||r|r|r|r|r|r|r|r|r||r|r|r|r|c|c|}
\hline
                   & \$/h & \multicolumn{4}{|c|}{Optimality Gap (\%)}  \\
Test Case & AC  & CP & NF+LL & SOC & SDP\\
\hline  
\hline  
Capacity & 985 & 89.84 & 88.93 & 88.93 & 1.32  \\
\hline 
Voltage & 102 & 1.96 & 1.96 & 1.96 & 0 \\
\hline 
\end{tabular}
\caption{AC-OPF Bounds using Relaxations on the 3-Bus Cases.}
\label{tbl:3_bus_results}
\end{table}



\section{Publicly Available Network Data}
\label{sec:networks}

Before proceeding further, it is important to survey and collect as many test cases as possible.  To the best of our knowledge, Table \ref{tbl:opf:inst} summarizes all of the readily available transmission system test cases.  Given that many of these cases were originally designed for testing AC Power Flow algorithms, special care is taken to highlight the source of (or lack of) generation capacity limits, generation cost functions, and line thermal limits.  Cells containing ``---" indicate missing data, which must be added before the test case will be suitable for optimization applications.\\

\begin{table}[t!]
\center
\begin{tabular}{|r||c|c|c|c|c|c|c|r|r|r|r|r|r|r|c|c|}
\hline
 & Original  & Generator  & Generator  & Thermal  \\
Name &  Source &  Capabilities &  Costs &  Limits \\
\hline
\hline
\multicolumn{5}{|c|}{IEEE Power Flow Test Cases}\\
\hline
14 Bus & \cite{IEEEBench} & --- & --- & --- \\
\hline
30 Bus & \cite{IEEEBench} & --- & --- & --- \\
\hline
57 Bus & \cite{IEEEBench} & --- & --- & --- \\
\hline
118 Bus & \cite{IEEEBench} & --- & --- & --- \\
\hline
300 Bus & \cite{IEEEBench} & --- & --- & --- \\
\hline
\hline
\multicolumn{5}{|c|}{IEEE Dynamic Test Cases}\\
\hline
17 Generator & \cite{IEEEBench} & --- & --- & --- \\
\hline
\hline
\multicolumn{5}{|c|}{IEEE Reliability Test Systems (RTS)}\\
\hline
RTS-79 & \cite{4113721} & \cite{4113721} & \cite{4113721,rts_79_gen_data} & \cite{4113721} \\
\hline
RTS-96 & \cite{780914} & \cite{4113721} & \cite{rts_79_gen_data} & \cite{4113721} \\
\hline
\hline
\multicolumn{5}{|c|}{Matpower Test Cases}\\
\hline
case2383wp & \cite{matpower} & \cite{matpower} & \cite{matpower} & \cite{matpower} \\
\hline
case2736sp & \cite{matpower} & \cite{matpower} & \cite{matpower} & \cite{matpower} \\
\hline
case2737sop & \cite{matpower} & \cite{matpower} & \cite{matpower} & \cite{matpower} \\
\hline
case2746wop & \cite{matpower} & \cite{matpower} & \cite{matpower} & \cite{matpower} \\
\hline
case2746wp & \cite{matpower} & \cite{matpower} & \cite{matpower} & \cite{matpower} \\
\hline
case3012wp & \cite{matpower} & \cite{matpower} & \cite{matpower} & --- \\
\hline
case3120sp & \cite{matpower} & \cite{matpower} & \cite{matpower} & ---\\
\hline
case3375wp & \cite{matpower} & --- & \cite{matpower} & --- \\
\hline
\hline
\multicolumn{5}{|c|}{PEGASE Test Cases}\\
\hline
case89pegase & \cite{rte_pegase} & \cite{rte_pegase} & --- &  \cite{rte_pegase}, partial \\
\hline
case1354pegase & \cite{rte_pegase} & \cite{rte_pegase} & --- & \cite{rte_pegase}, partial \\
\hline
case2869pegase & \cite{rte_pegase} & \cite{rte_pegase} & --- & \cite{rte_pegase}, partial \\
\hline
case9241pegase & \cite{rte_pegase} & \cite{rte_pegase} & --- & \cite{rte_pegase}, partial \\
\hline
\hline
\multicolumn{5}{|c|}{Edinburgh Test Case Archive}\\
\hline
Iceland & \cite{EDINBench} & --- & --- & --- \\
\hline
Great Britain &  \cite{EDINBench} &  \cite{EDINBench}& --- & --- \\
\hline
Reduced G.B. &  \cite{EDINBench} &  \cite{EDINBench} & --- &  \cite{EDINBench} \\
\hline
%
\hline
\multicolumn{5}{|c|}{EIRGrid Test Cases}\\
\hline
Summer Valley 2013 & \cite{EIRGrid} & \cite{EIRGrid} & --- & \cite{EIRGrid} \\
\hline
Summer Peak 2013 & \cite{EIRGrid} & \cite{EIRGrid} & --- & \cite{EIRGrid} \\
\hline
Winter Peak 2013-14 & \cite{EIRGrid} & \cite{EIRGrid} & --- & \cite{EIRGrid} \\
\hline
%
%
\hline
\multicolumn{5}{|c|}{Publication Test Cases}\\
\hline
3-Bus& \cite{6120344} & \cite{6120344}  & \cite{6120344}  & \cite{6120344}  \\
\hline
case4gs & \cite{grainger1994power} & \cite{grainger1994power}  & ---  & --- \\
\hline
case5 & \cite{5589973} & \cite{5589973} & \cite{5589973}  & --- \\
\hline
case6-ww & \cite{wollenberg2006power} & \cite{wollenberg2006power}  & \cite{wollenberg2006power}  & \cite{wollenberg2006power}  \\
\hline
case6bus & \cite{crow2002computational} & ---  & ---  & --- \\
\hline
%
case9 & \cite{Dembart_Erisman_Cate_Epton_Dommel_1977} & \multicolumn{3}{c|}{\cite{matpower}, original source unknown} \\
\hline
case30-as & \cite{4075418} & \cite{4075418} & \cite{4075418} & \cite{4075418} \\
\hline
case30-fsr & \cite{4075418} & \cite{630479} & \cite{630479} & \cite{4075418} \\
\hline
case39 & \cite{Bills_1970} & \cite{Bills_1970, pai1989energy} & \cite{4562306} & \cite{TCCalculator} \\
\hline
\end{tabular}
\caption{Survey of Transmission System Data Sources}
\label{tbl:opf:inst}
\end{table}

\noindent
In conducting this survey, some test cases were modified or omitted and a number of useful observations were made.  The rationale for these modifications and omissions is as follows.
\subsection{Network Modifications}

\paragraph{Small \matpower Test Systems}
\matpower mostly contains test systems with sufficient data for AC-OPF studies.  However, there are a few small cases with only enough data for AC power flow studies, namely, case4gs and case6bus.  The following minor modifications are made to prepare these for AC-OPF studies.  
\begin{enumerate}
\item case4gs -- Generator costs are assigned similarly to the ``cdf2mp'' algorithm provided with \matpower, Pmax increased to 300 MW on the generator at bus 1, Qmax and Qmin are increased to 150,-150 MVar respectively to make the model feasible.
\item case6bus -- Generator costs are assigned similarly to the ``cdf2mp'' algorithm.
\end{enumerate}

\paragraph{Matpower 30 Bus Test System}
This test case was originally defined in \cite{4075418} and latter modified in \cite{630479}, with the later version being distributed with \matpower.  The generator capacities and cost functions are significantly different in these two cases, hence we choose to include both using the identifiers {\em as} and {\em fsr} to differentiate the two cases.\footnote{The identifiers are based on the first initial of the authors' last names.}

\paragraph{EIRGrid Networks}
These test cases are provided in PSS/E--30 format and need to be converted into \matpower case files.  The PSS/E file conversion tools provided with Pylon \cite{pylon_sim} and \matpower did not lead to converging AC power flows test cases on these networks.  This may be caused in part to the explicit $\Delta$Y thirty degree phase shifts which are typically omitted from AC power flow studies \cite{9780070359581} and several three winding transformers with negative resistance values.  To address these challenges, we developed our own PSS/E--30 to \matpower translation tool, which produces converging AC power flow cases for these systems.  However, these translations are only preliminary as we did not have access to PSS/E for verification of the translation.  Detailed documentation of the translation process appears in the test case files.

\subsection{Network Omissions}
%
%
\paragraph{IEEE 30 Bus ``New England" Dynamic Test System}
This test case is nearly identical to the IEEE 30 test case and would not bring additional value to this archive.

\paragraph{IEEE 50 Generator Dynamic Test System}
In its specified state, this test case does not converge to an AC power flow solution.  However, if the active generation upper bounds are increased to 1.5 times their given value, and voltage bounds are set to $1 \pm 0.16$ V p.u., then a solution can be obtained.  This solution still exhibits significant voltage drops, atypical of other networks.  Many of the lines in this network have negative $r$ and $x$ values, and it is unclear why. Also the size of the generating units are one or two orders of magnitude larger than any documented generation unit in the U.S\@.  Due to these concerns, this network is omitted. 

\paragraph{Networks for Optimal Transmission Switching}
A number of variants of the IEEE 118 \cite{Blumsack06} and IEEE RTS-96 \cite{5589434} networks have been developed for use in optimal transmission switching studies.  However, due to the difficulties with AC feasibility \cite{pscc_ots} we choose not to include them.

\paragraph{Local Minima Test Case Archive}
The University of Edinburgh, School of Mathematics has developed an archive of tests cases for the AC-OPF with known local minima \cite{6581918,local_opt_archive}.  These test cases are very useful for testing AC-OPF algorithms, however, they are often significant departures from real-world inputs.  In the interest of keeping this archive based on realistic test cases, we omit this local minima archive.\\


\subsection{Network Observations}

\paragraph{IEEE 300 Test System}
This test case exhibits very large voltage drops, with a voltage drop across branch 191--192 of 0.09 volts p.u.\ and 20 degrees. Such drops are not observed in the larger real-world networks \cite{Purchala:2005gt}, which brings the realism of this network into question.  Also note that in the \matpower version of this network a phase shifting transformer is missing between buses 196 and 2040.

\paragraph{IEEE RTS-96}
Note that RTS-96 \cite{780914} is simply three RTS-79 \cite{4113721} networks connected together.  One of the subtle points in these networks is the line charging on line $106$--$110$, which is 2.459 p.u\@.  In RTS-79, a -1.0 p.u.\ bus-shunt is used at bus 6 to balance out this massive line charge.  This value must be used in all three analogous parts of the RTS-96 network to produce a feasible AC power flow.  The generation costs presented in \cite{780914,4113721} are not provided in \$/h, hence we adopt to interpretation of \cite{rts_79_gen_data} to get them in the correct units.

\paragraph{3000 Bus Polish Test Cases}
Unlike the other Polish network cases, these test cases have 10 lines with negative resistance values.  Hence, these lines can inject power into the network and invalidate some models, such as the AC copper plate relaxation.  However, these negative values do not appear to adversely effect the  power flow models considered in this report. 

\paragraph{PEGASE Test Cases}
The Pan European Grid Advanced Simulation and State Estimation (PEGASE) Test Cases where originally studied in \cite{6488772} and are now distributed with \matpower and on arXiv \cite{rte_pegase}.  The PEGASE cases with less than 9000 buses reflect an aggregated version of the Pan European Grid and hence have parameters that do not reflect individual network components (e.g. they include generators that can consume active power).
\\
\\
\noindent
As indicated by  Table \ref{tbl:opf:inst}, many of these networks are missing key information relating to generator capabilities, costs, and line thermal limits.  We now turn to developing models to fill in the gaps in these test cases. 


\section{Data Driven Models}
\label{sec:data_models}

In an ideal situation, the data missing from Table \ref{tbl:opf:inst} would be incorporated by returning to the original network design documents and extracting key specification information, such as generator nameplates and conductor specification sheets.  Unfortunately, due to age of many of these test cases, the original sources are often unknown or inaccessible.  A common approach for developing the missing data is to adopt simple equations, as exemplified by \matpower\cite{matpower}.  Drawing inspiration from the big data movement, we instead take a statistical approach built on publicly available datasets.  Such data driven models may not reflect any specific real world network, but at least they reflect many of the statistical features found in real networks.  As identified in Section \ref{sec:networks}, the key pieces of missing data are generator capability curves, generation costs, and line thermal limits.  Throughout the rest of this section we analyze publicly available data sets, and propose data driven models that can fill the gaps in these test cases.   


\subsection{Generation Models}

In classic AC power flow test cases, very little information is given about each generation unit.  Typically, only the active power generation at one point in time and the reactive generation limits are provided.  In contrast, optimization applications are best served by detailed models of the each generator's specification, including active/reactive power capability curves, ramp rates, startup/shutdown costs, and fuel efficiency.  The challenge in building these generator models is to extrapolate reasonable operating capabilities from just one snapshot in time, that is, the operating point of the generator in the AC power flow case.  To achieve this task, we first observe that the bulk of a generator's properties are driven by its mechanical design, which is in turn significantly influenced by its fuel type.  Hence, we begin developing a data driven model for generators by assigning them a fuel category.  Once a fuel category is determined, probabilistic models for both the active power capabilities and fuel costs of generators can be derived from publicly available data sources.

\subsubsection{Reviewing the Data}

The U.S.\ Energy Information Administration (EIA) collects vast amounts of data on generation units throughout the U.S\@.  Two reports are particularly useful to this study, the detailed generator data (EIA-860 2012 \cite{EIA860}) and state fuel cost data (SEDS \cite{SEDS2012}).  The fuel costs data is sufficiently summarized in the table, ``Primary Energy, Electricity, and Total Energy Price Estimates, 2012" within SEDS\@.  However, the generator data requires some processing before it can be used here.  

Form EIA-860 provides detailed metadata for 19,000 generation units in the U.S\@.  The key fields of for this study are status (e.g., operational, out of service, standby), energy source (i.e. the primary fuel type), Nameplate Capacity (MW), and Summer Nameplate Capacity (MW).  We focus our attention on generators that are currently in service and have a single fuel type.  This reduces the total number of generating units under consideration to 12,000.  The EIA-860 form classifies generation fuel categories into detailed classifications, e.g., Residual Fuel Oil, Synthetic Gas, and Black Liquor (for a complete list see, \cite{EIACodes}).  We group these detailed classifications into the following broad categories: Coal (COW), Petroleum (PEL), Natural Gas (NG), Biomass (BIO), Nuclear (NUC), Dispatchable Renewables (DRN), and Renewables (RN).  A complete mapping of fuel category codes is provided in Table \ref{tbl:gen:fuel:map}. 
\begin{table}[h!]
\center
\begin{tabular}{| l | l | l | }
\hline
Fuel Category & Category Label & EIA-860 Labels \\
\hline
\hline
Coal & COW & ANT,BIT,LIG,SUB,WC \\
\hline
Petroleum & PEL & RC, DFO, JF, KER, PC, RFO, WO  \\
\hline
 Natural Gas & NG & BFG, NG, OG, PG, SG, SGC  \\
\hline
Nuclear & NUC & NUC \\
\hline
Biomass & BIO & AB,MSW,OBS,WDS,OBL,SLW,BLQ,WDL  \\
\hline
Dispatchable Renewables & DRN & WAT,GEO \\
\hline
Renewables & RN & SUN, WND\\
\hline
\end{tabular}
\caption{Generator Fuel Category Mapping}
\label{tbl:gen:fuel:map}
\end{table}
Before building the generation models, we elect to omit the Biomass, Renewables, and Dispatchable Renewables fuel categories for the following reasons.
\begin{enumerate}
\item RN -- There is an implicit assumption in classic power system optimization problems that the generators are actively controlled via voltage magnitude and active power set-points.  Hence, incorporating non-dispatchable generation is outside the scope of these problems.
\item DRN --  To the extent that generation costs are modeled in this report, the marginal cost for fuel in renewable resources is considered to be 0.  This may lead to a degenerate behavior in optimization problems studying nodal marginal prices.  Dispatchable Renewables are omitted to avoid this degeneracy issue.
\item BIO --  The generators within the biomass category are statistically quite similar to those in the petroleum category, and hence do not bring much additional value.  This category is omitted in the interest of building a concise collection of fuel categories.
\end{enumerate}
After filtering the fuel categories, 7,800 generation units remain from the previous set of 12,000.  The following generator models are built using this filtered version of the EIA 2012 dataset.

\subsubsection{Generation Fuel Category Classification Model}

\begin{figure}[h!]
\center
    \includegraphics[height=6.5cm]{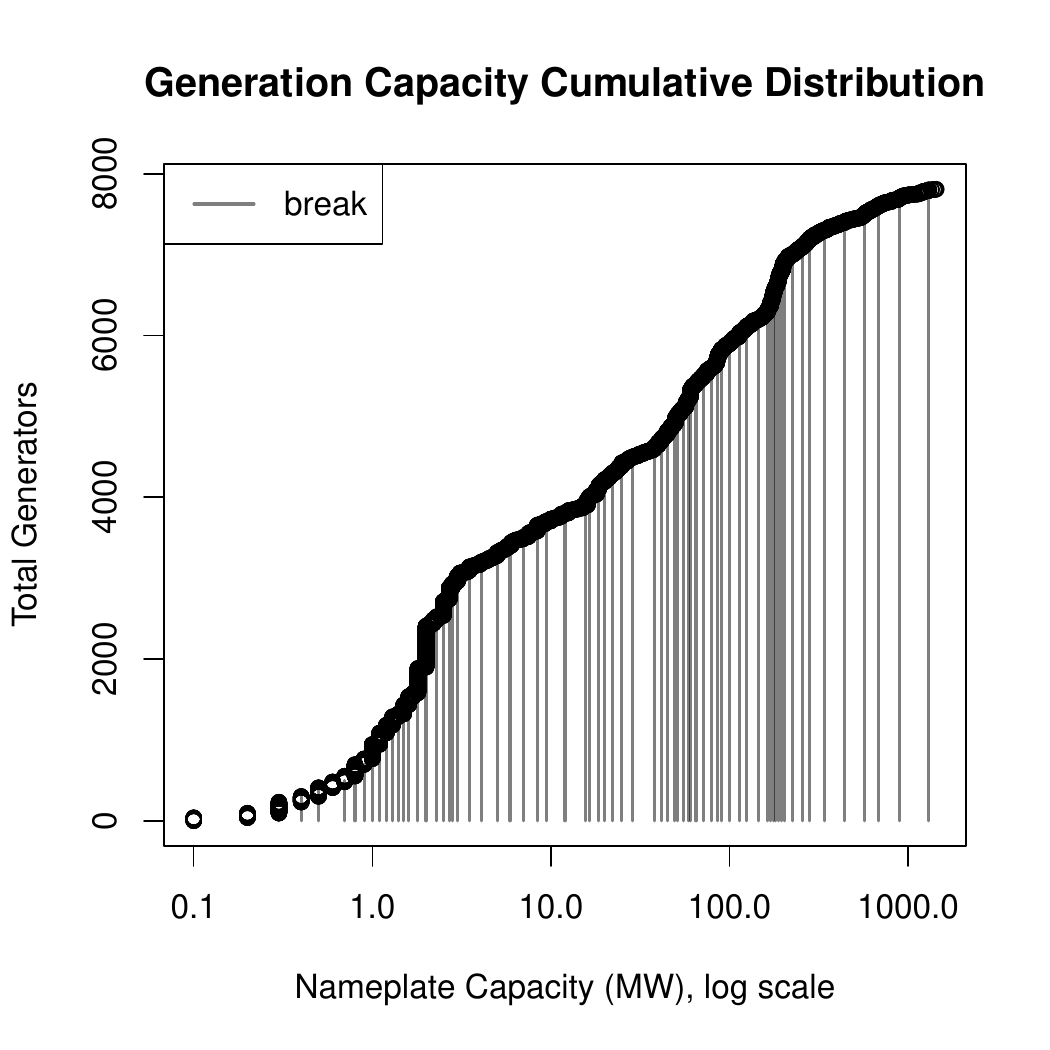}
    \includegraphics[height=6.5cm]{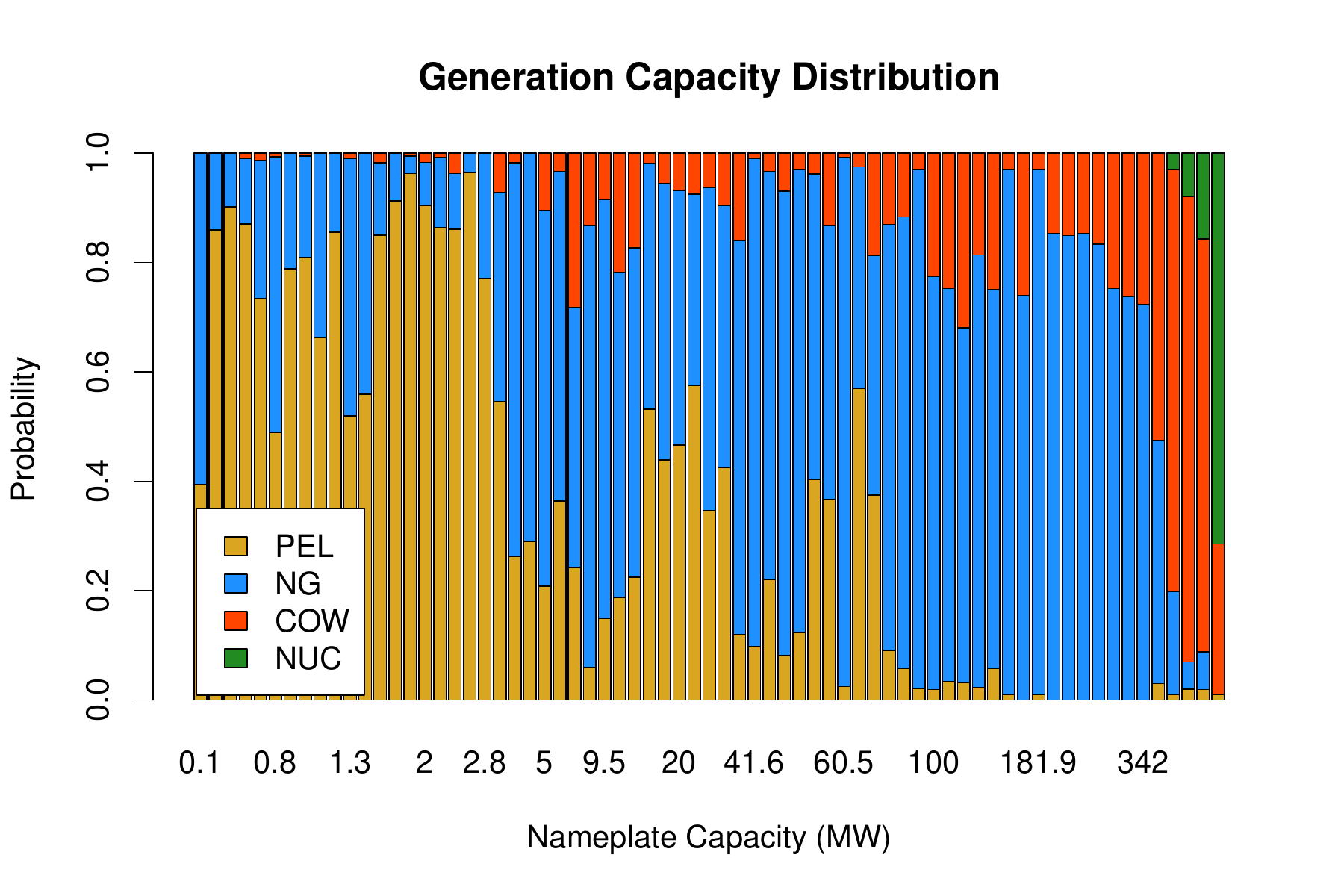}
    \caption{Distribution of Generator Nameplate Capacities (left), Empirical Distribution of Generation Fuel Categories (right).}
\label{fig:gen_cap_bins}
\end{figure}

The first step in developing the generation unit models is fuel classification.  Given that we have thousands of data points for generation units, it is possible to build an accurate empirical distribution of the data.  We begin by sorting the generation units by their nameplate capacity. Observing that the nameplate capacities are not linearly distributed, we then break the data into contiguous bins containing at least 100 samples and build an empirical distribution from these bins, as in Figure \ref{fig:gen_cap_bins}. A fuel category classifier is easily built by selecting the corresponding nameplate capacity bin and rolling a weighted die to select the fuel type.  When available, the fuel category classifier is applied on the maximum active generation value provided with the test case.  However, when a maximum active generation value is unavailable, the current active generation value is a suitable alternative.
 
%

One important point, which is easily overlooked, is the identification of synchronous condensers.  In classic power flow studies such devices are not explicitly identified, and are modeled as voltage controlled generators with no active generation.  To identify such devices, we introduce a new fuel category called SYNC, and any generator with an active generation upper bound of 0 is assigned this value.  There is also one special case to be considered in generator category classification.  Because many of these networks were originally designed as AC power flow test cases, the generators at the slack bus sometimes have no active generation value specified, as this would be a free variable in a power flow study.  If this occurs, we assume the generator is a very large and cheap power plant and assign it the class NUC\@.  The empirical distribution shown in Figure \ref{fig:gen_cap_bins} combined with these special cases forms the generation fuel classification model (GF-Stat).

\subsubsection{Generation Capacity Models}

An ideal generation capacity model would produce a complete generation capacity curve (e.g., Figure 15.8 from \cite{9780070359581}).  However, power system optimization models often approximate the generation capacity curve as a box by specifying upper and lower bounds on active and reactive power injection.  We adopt this approximation and focus first on determining a reasonable upper limit on active power capabilities.

\paragraph{Active Generation Capability}
Before building statistical models for active generation capacities, the EIA data requires further filtering to be aligned with the existing test cases.  Specifically, the EIA data contains many generators with less than 1MW nameplate capacity, in contrast over 90\% of the generators in the test cases are above 1MW.  To best capture the generators within in the available test cases, we remove generators below 5 MW from the EIA dataset resulting in a set of 4,500 units.  The differences between these various distributions are summarized in Table \ref{tbl:gen:gen:stat}.

\begin{table}[h!]
\center
\begin{tabular}{|l||r||r|r|r|r|r|r|r|r|r|r|r|r|r|r|c|c|}
\hline
& & \multicolumn{7}{c|}{Quantiles, Capacity (MW)}\\
& $n$ & Min & 10\% & 25\% & Median & 75\% & 90\% & Max \\
\hline
\hline
Test Case, Active Generation & $910$ & 0.1 & 2.4 & 15.8 & 76 & 271 & 313 & 2520  \\
\hline
\hline
EIA, Nameplate Capacity & $7808$ & 0.1 & 1.0 & 2.0 & 16 & 94 & 233 & 1440  \\
\hline
EIA, Nameplate Capacity ($\geq 5$MW) & $4531$ & 5.0 & 10.0 & 25.0 & 75 & 188 & 390 & 1440  \\
\hline
\end{tabular}
\caption{Active Generation and Nameplate Capacity Distributions}
\label{tbl:gen:gen:stat}
\end{table}

\begin{figure}[h!]
\center
 \includegraphics[width=6.5cm]{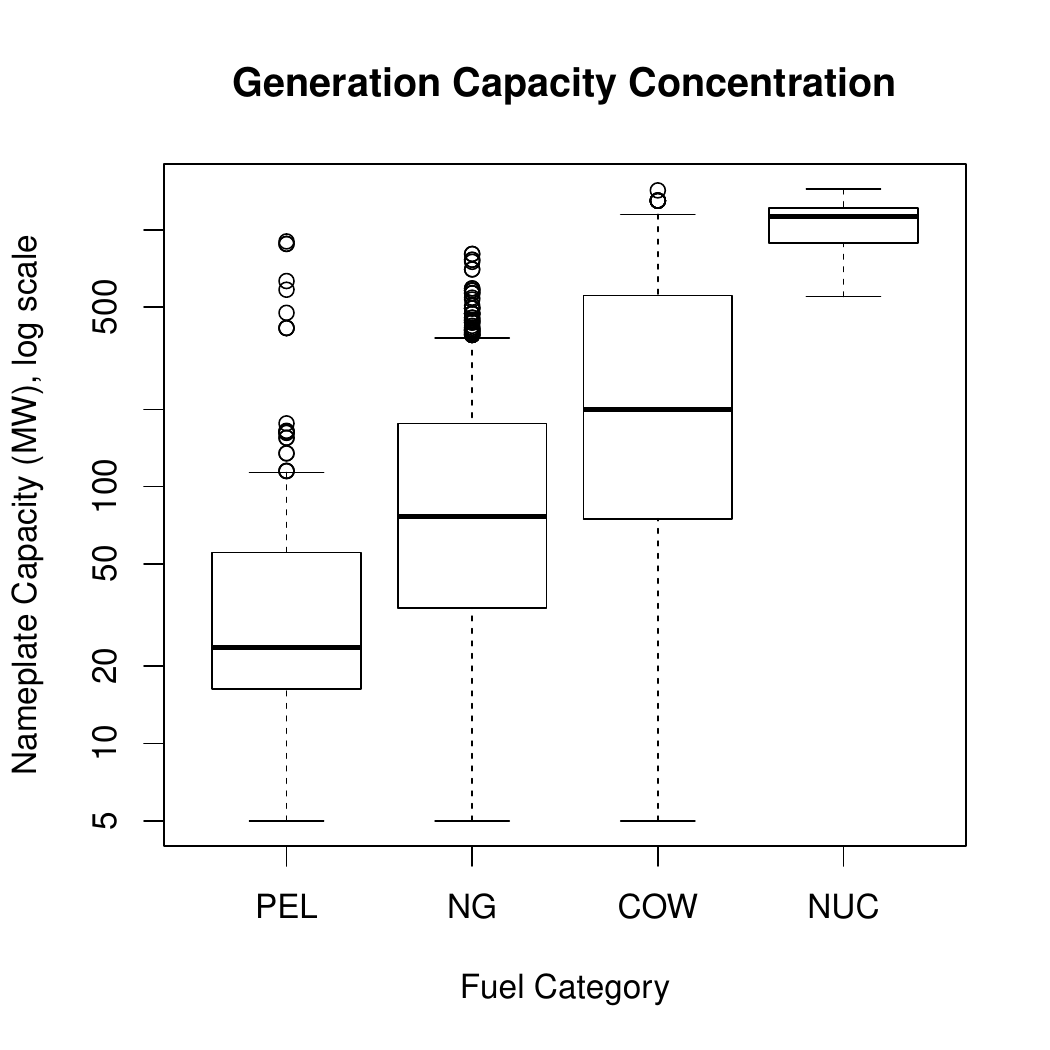}
    \caption{Distribution of Nameplate Capacities for each Fuel Category.}
\label{fig:gen_cap_fuel_type}
\end{figure}

\begin{figure}[h!]
\center
    \includegraphics[height=6.5cm]{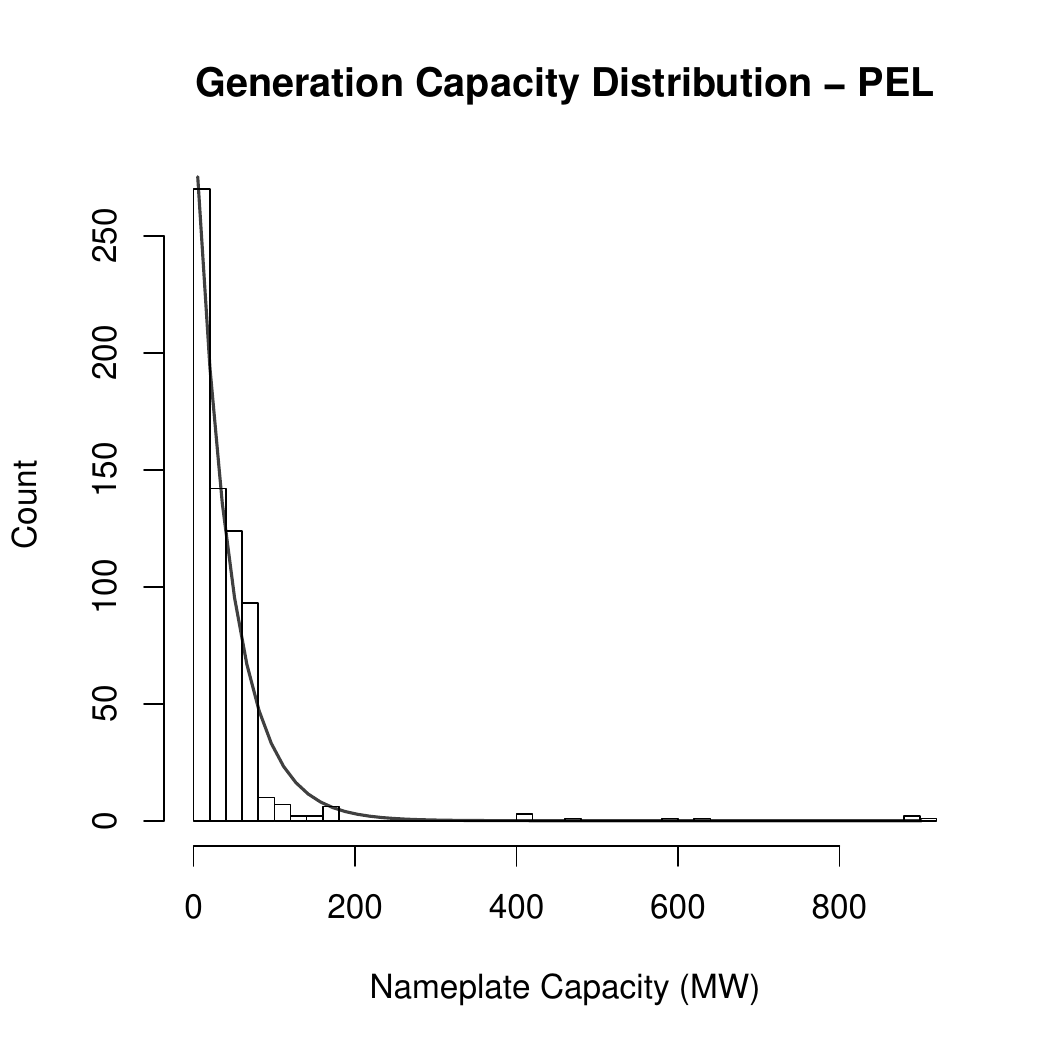}
    \includegraphics[height=6.5cm]{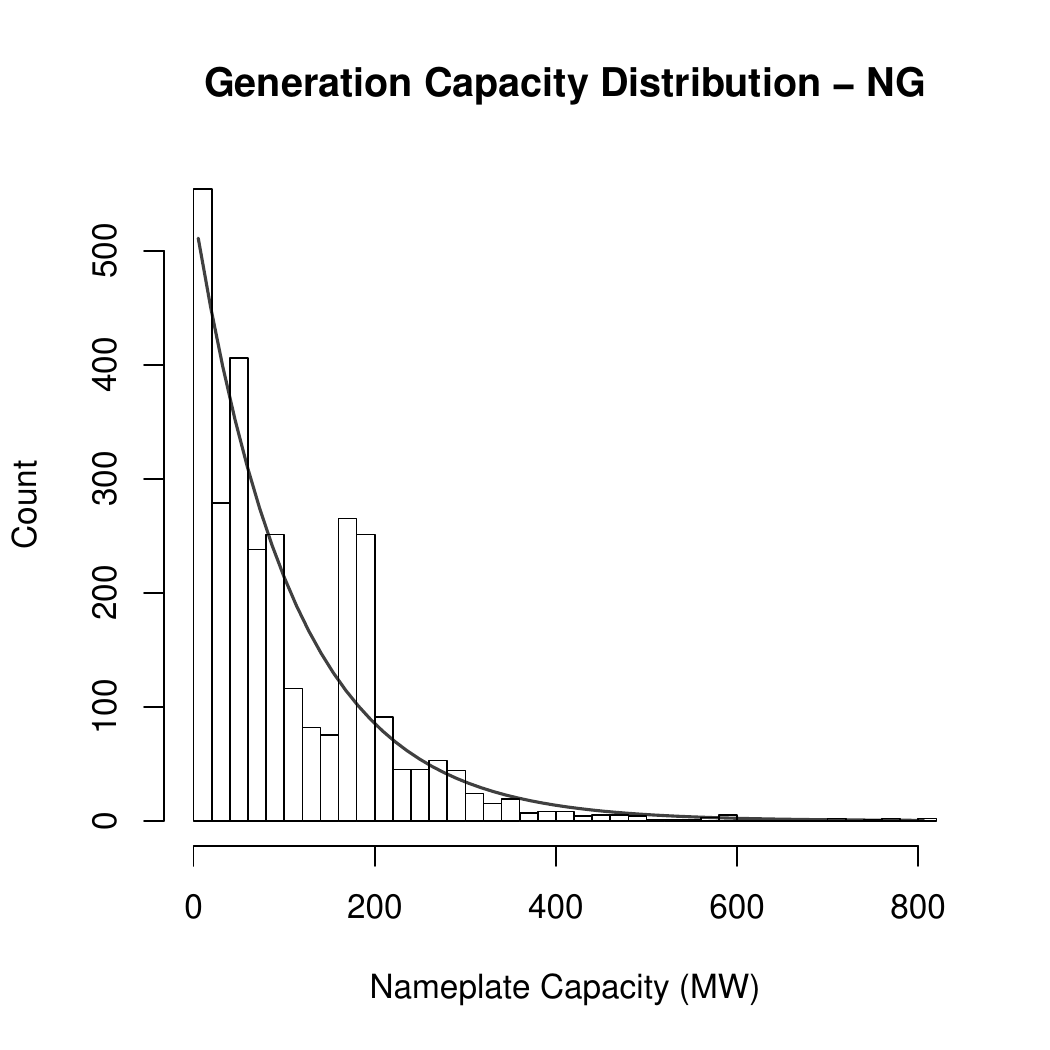}
    \includegraphics[height=6.5cm]{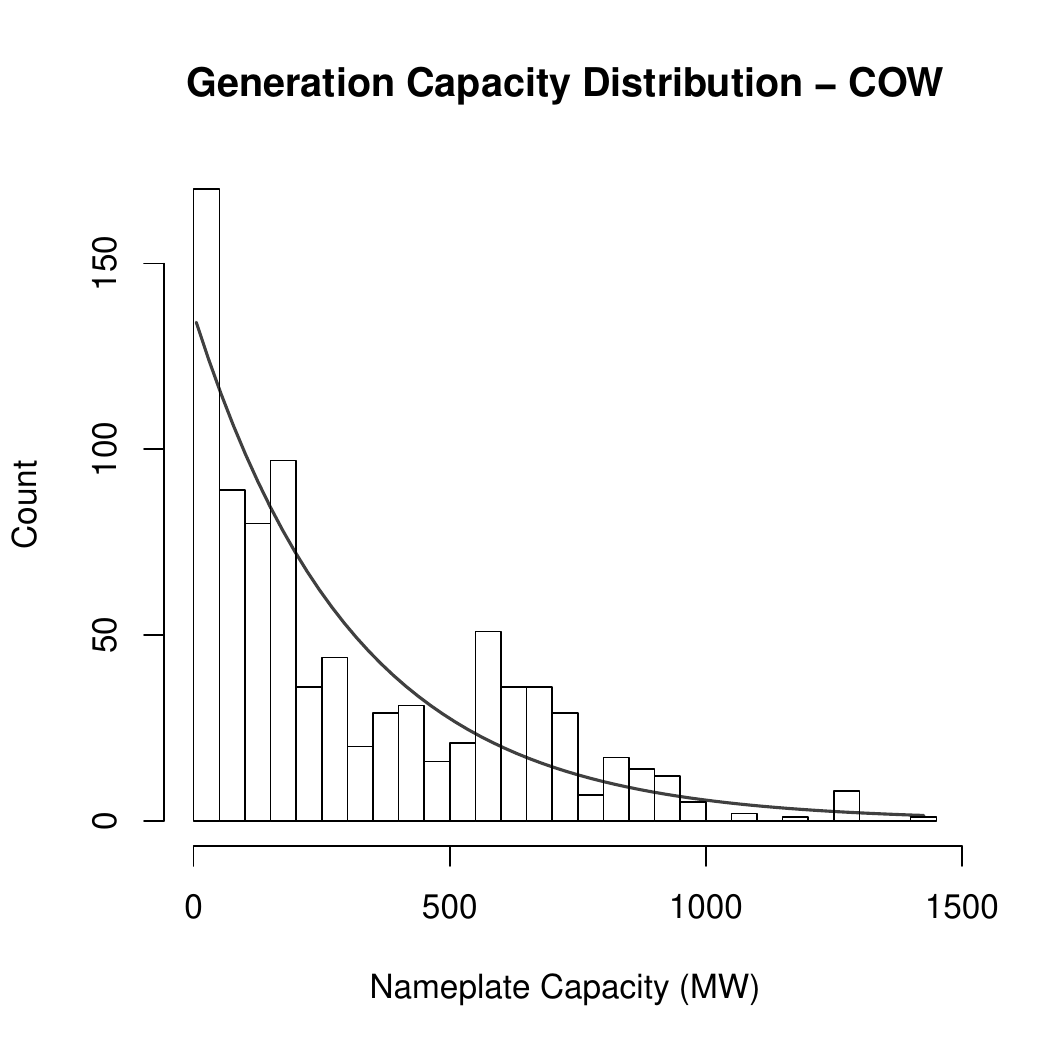}
    \includegraphics[height=6.5cm]{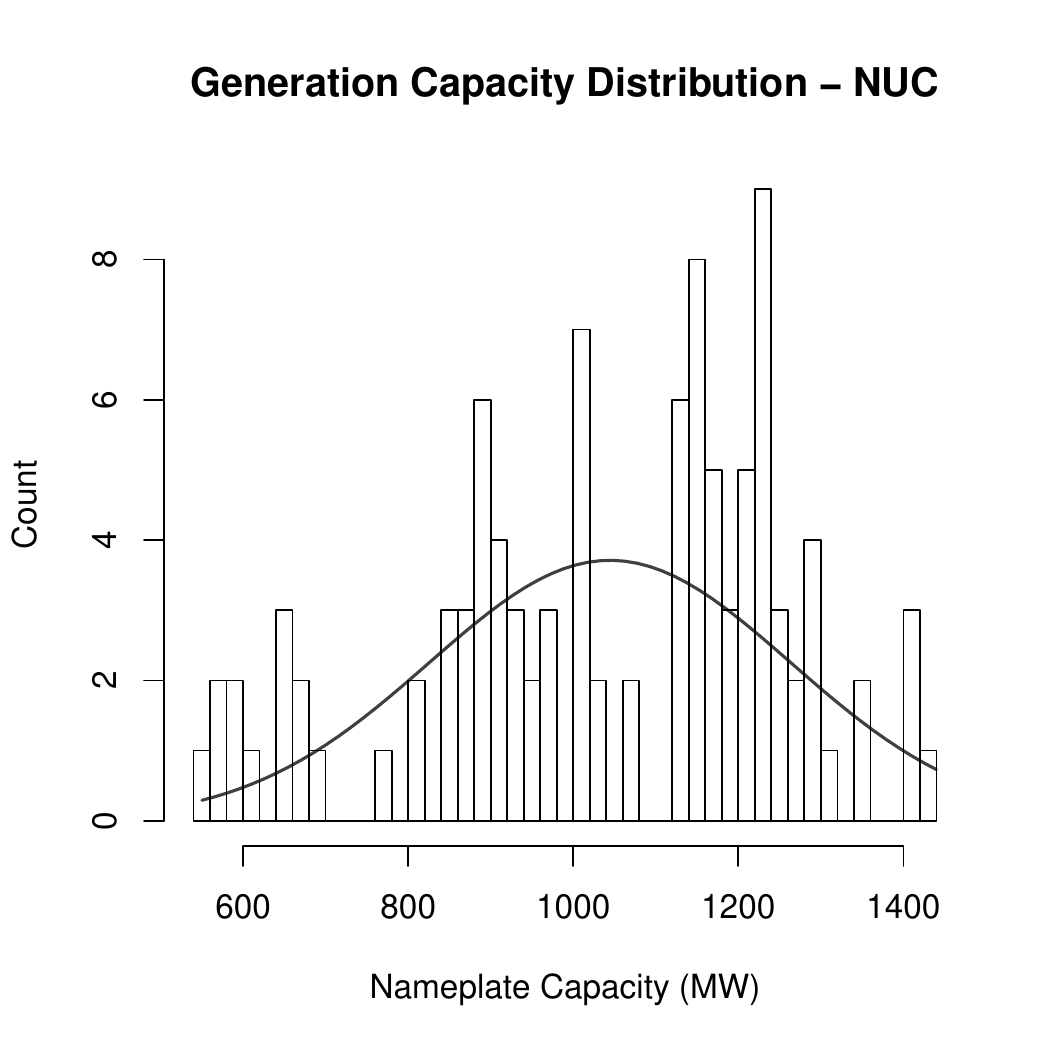}
    \caption{Distribution of Generator Nameplate Capacities with Probability Distributions.}
\label{tbl:gen:cap:stat:dist}
\end{figure}

\begin{figure}[h!]
\center
 \includegraphics[width=6.5cm]{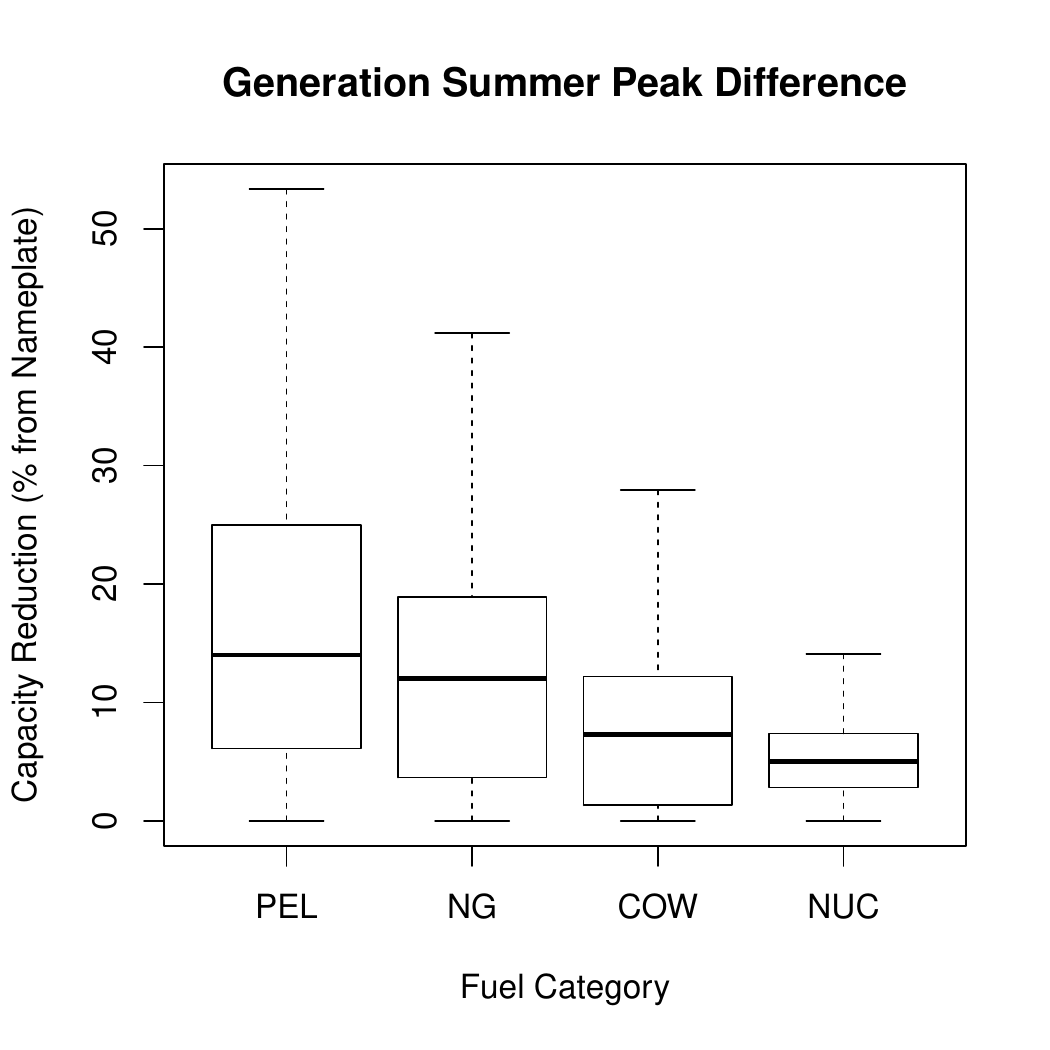}
    \caption{Distribution of Summer Peak Generation Capacity Reduction by Fuel Category.}
\label{tbl:gen:cap:peak:dist}
\end{figure}

Grouping the active power nameplate capacity by fuel type, Figure \ref{fig:gen_cap_fuel_type} and Table \ref{tbl:gen:cap:stat} illustrate the clear differences across fuel categories.  Notably, although each fuel category has a wide range of nameplate capacities, the medians of each fuel category are significantly different.  Figure \ref{tbl:gen:cap:stat:dist} further investigates the distributions within each fuel type and reveals that an exponential distribution is a suitable model for PEL, NG, and COW, while a normal distribution is suitable for NUC\@.  The parameters from maximum likelihood estimation of these distributions are presented in Table \ref{tbl:gen:cap:model}.  We also consider the the summer peak capacity reduction of each fuel category, as in Figure \ref{tbl:gen:cap:peak:dist} and Table \ref{tbl:gen:cap:model}, to build a simple linear model for extreme cases when the more detailed probabilistic models fail.

\begin{table}[h!]
\center
\begin{tabular}{|l||r||r|r|r|r|r|r|r|r|r|r|r|r|r|c|c|}
\hline
                        &        & \multicolumn{7}{c|}{Quantiles, Capacity (MW)}\\
Fuel Category & $n$ & Min & 10\% & 25\% & Median & 75\% & 90\% & Max \\
\hline
\hline
PEL & 665 & 5 & 8 & 16 & 24 & 55 & 67 & 902 \\
\hline
NG & 2912 & 5 & 10 & 34 & 77 & 176 & 227 & 806 \\
\hline
COW & 852 & 5 & 12 & 75 & 200 & 556 & 720 & 1426 \\
\hline
NUC & 102 & 550 & 680 & 889 & 1127 & 1216 & 1289 & 1440 \\
\hline
\end{tabular}
\caption{Generator Nameplate Capacity Statistics by Fuel Category.}
\label{tbl:gen:cap:stat}
\end{table}

\begin{table}[h!]
\center
\begin{tabular}{|l||r||r|r||r|r||r|r|r|r|r|r|r|r|r|c|c|}
\hline
                   & \multicolumn{2}{c|}{Nameplate Capacity (MW)} \\
 Fuel  Type & $\hat{\lambda}$ & Summer Reduc.  $\hat{\mu}$ \\
\hline
\hline
PEL & 0.023254 & 16.11\% \\
\hline
NG & 0.009188 & 12.98\% \\
\hline
COW & 0.003201 & 8.48\% \\
\hline
\end{tabular}

\begin{tabular}{|l||r|r||r|r|r||r|r|r|r|r|r|r|r|r|c|c|}
\hline
                   & \multicolumn{3}{c|}{Nameplate Capacity (MW)} \\
 Fuel  Type & $\hat{\mu}$ & $\hat{\sigma}$ & Summer Reduc.  $\hat{\mu}$ \\
\hline
\hline
NUC & 1044.56 & 219.27 & 5.80\% \\
\hline
\end{tabular}

\caption{Distribution Parameters for Generator Capacity Models.}
\label{tbl:gen:cap:model}
\end{table}

Combining all of these models, an active generation capacity model (AG-Stat) is constructed as follows.  Given a fuel category $f$ and an active generation upper limit or present output $p^g$, the fuel category nameplate capacity distribution is sampled as $p^{gu}$, until $p^{gu} > p^g$.  The maximum active generation is then assigned the value of $p^{gu}$.  However, if 100 samples are attempted and none satisfy $p^{gu} > p^g$, a last resort linear model based on summer peak reduction is used.  It is assumed that $p^g$ is the value of the generator running at the summer peak capacity and the maximum active generation is assigned as $p^g$ increased to the appropriate nameplate capacity, as specified in Table  \ref{tbl:gen:cap:model}.


\paragraph{Reactive Generation Capabilities}

In synchronous machines, reactive generation capabilities are tightly coupled with active generation capabilities.  Lacking detailed information about the generator's specifications and having put significant effort into an active power capability model, simple models for reactive generation capabilities are developed.  Using ``Reactive Capability Curves of a Hydrogen Cooled Generator at Rated Voltage", Figure 15.8, \cite{9780070359581} as a guide, it is observed that the reactive power capability of a synchronous machine is roughly $\pm 50\%$ of its nameplate capacity.  This inspires two reactive power models,
\begin{enumerate}
\item At Most 50 (RG-AM50) -- This model assumes the given reactive power bounds are accurate, unless they exceed 50\% of the nameplate capacity, in which case they are reduced to $\pm 50\%$ of the nameplate capacity.  This provides a pessimistic model of the generator's capabilities.
\item At Least 50 (RG-AL50) -- This model assumes the given reactive power bounds are accurate, unless they are below 50\% of the nameplate capacity, in which case they are increased to $\pm 50\%$ of the nameplate capacity.  This provides an optimistic model of the generator's capabilities.
\end{enumerate}
These models are not often needed in this report, as the majority of cases studies come with reasonable reactive power bounds.  However, such models are essential when developing significant departures from the standard test cases.  For example, when developing test cases for long-term network expansion studies, where the reactive power capabilities can play an important role in the overall network design \cite{pscc_ep}.

\subsubsection{Generation Cost Model}

Modeling generation costs could easily become a quagmire combining capital investment, game theory, agent modeling, and complex regulatory structures.  However, like many classic power system optimization problems (e.g., AC-OPF), we choose to avoid these modeling challenges by simply focusing on the marginal costs of power generation in a non-competitive environment.  Namely, the entire network strives to minimize its cumulative fuel costs.  We obtain fuel price information from ``Primary Energy, Electricity, and Total Energy Price Estimates, 2012" in the SEDS dataset \cite{SEDS2012}.  The SEDS fuel cost data is broken down by state and provided in Dollars per Million Btu. The conversion factor, 1 Million btu $=$ 0.29307107 MWh, is used to convert these costs into appropriate units for a power system optimization problem.  We observe that, for the fuel categories of interest, the fuel costs are roughly normally distributed, with the parameters specified in Table \ref{tbl:gen:cost:model}.  The model for generation costs (AC-Stat) is built as follows.  We assume that the fuel cost parameters from Table \ref{tbl:gen:cost:model} are representative of the price variations across the various generating units, so that, given a fuel category, we simply draw a sample from the associated normal distribution to produce a fuel cost value.  This value becomes the \$/MWh cost of that generator in the power system optimization problem.

\begin{table}[h!]
\center
\begin{tabular}{|r|r||r|r|r|r||r|r||r|r|r|r|r|r|r|c|c|}
\hline
& & \multicolumn{2}{c|}{Cost (\$/MWh)} \\
Fuel Category & SEDS Label & $\hat{\mu}$ & $\hat{\sigma}$ \\
\hline
\hline
PEL & Distillate Fuel Oil & 6.8828 & 0.3334 \\
\hline
NG & Natural Gas & 1.0606 & 0.2006 \\
\hline
COW & Coal & 0.7683 & 0.2452 \\
\hline
NUC & Nuclear Fuel & 0.2101 & 0.0199 \\
\hline
\end{tabular}
\caption{Generator Cost Model.}
\label{tbl:gen:cost:model}
\end{table}

At this point we have developed models for all of the key generation parameters, namely, active/reactive generation capabilities and generation costs.  We now turn our attention to a transmission line thermal limit model, to complete the missing information for power system optimization.

\subsection{Thermal Limit Models}

Determining a transmission line's thermal rating is fairly straightforward when the conductor type and line length are available.  Unfortunately, in all of the test cases studied here, none of this information is available.  The only available line parameters are the impedance ($r+ i x$ p.u.), line charge ($b^c$ p.u.), and often the nominal voltage ($\dot{v}$) on the connecting buses.  Given this limited information, it is common to estimate the thermal limit with a Surge Impedance Loading (SIL) \cite{9780070359581,pes_opf_comp}.  However, a significant shortcoming of this approach is that all of the lines within a given voltage level have the same thermal limit.  This does not reflect real-world conditions, where lines have a wide range of thermal limits within each voltage level.  This distinction is important for optimization applications, as it only takes one line with a low thermal limit to cause significant congestion in the power flow.  In the rest of this section, a simple linear regression model is developed around the values of $r$, $x$ and $\dot{v}$ to go beyond SIL\@.  Additionally, a reasonable upper bound for the line thermal limit is derived for cases where any of the $r$, $x$ and $\dot{v}$ values are available.

\subsubsection{Reviewing the Data}

\begin{figure}[h!]
\center
    \includegraphics[width=6.5cm]{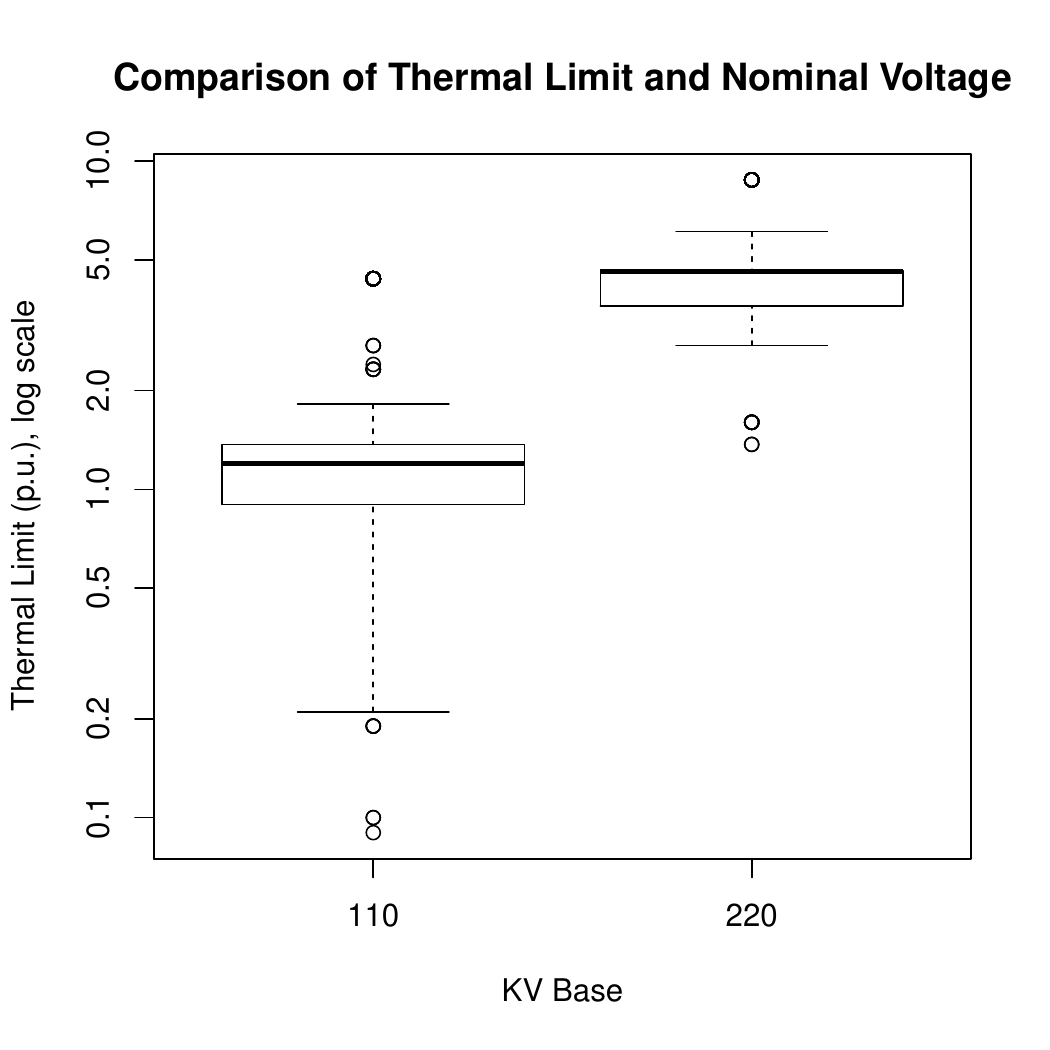}
    \includegraphics[width=6.5cm]{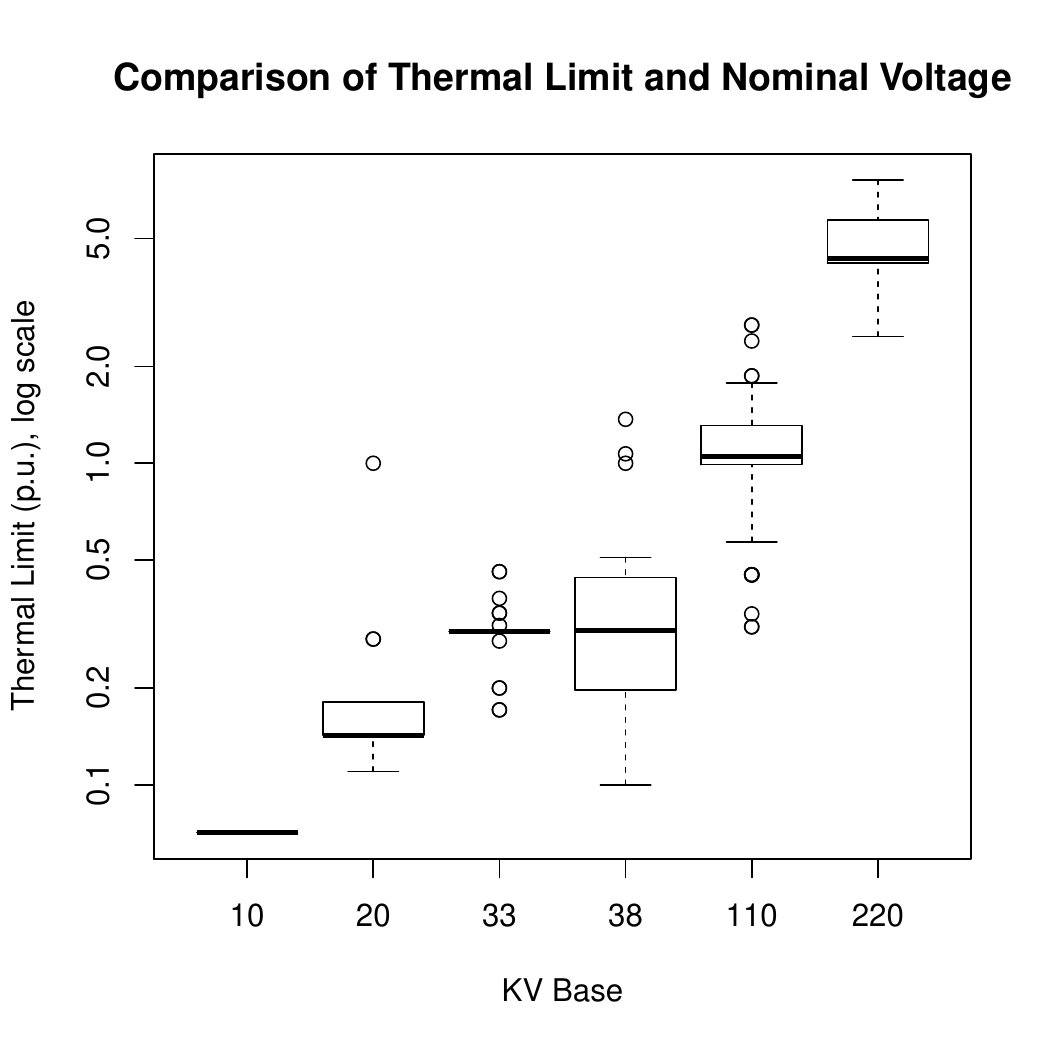} 
    \caption{Polish Network case2383wp (left) and EIRGrid Summer Night Valley 2013 (right).}
\label{fig:cap_vs_kv}
\end{figure}

Realistic line thermal limits are available from two sources, the Polish transmission system data provided with \matpower \cite{matpower} and the Irish transmission system data provided by EIRGrid \cite{EIRGrid}.   It was observed that these data sets do include a few lines without thermal limits, represented by large possessive numbers.  These lines consisted primarily of transformers and lines below 20kV.  To focus on meaningful data, we only consider lines that are not transformers, with thermal limits below 9.8 p.u., and nominal voltages below 250kV\@.  After filtering, Figure \ref{fig:cap_vs_kv} shows the line capacities of these networks broken down by kV Base.  The mean capacity within each nominal voltage group clearly supports the SIL approach.  However, the significant variance, which even leads to overlap between neighboring voltage levels, reaffirms the point that in real networks both low and high capacity lines exist for a given voltage.

\subsubsection{A Statistical Model}

\begin{figure}[h!]
\center
    \includegraphics[width=6.5cm]{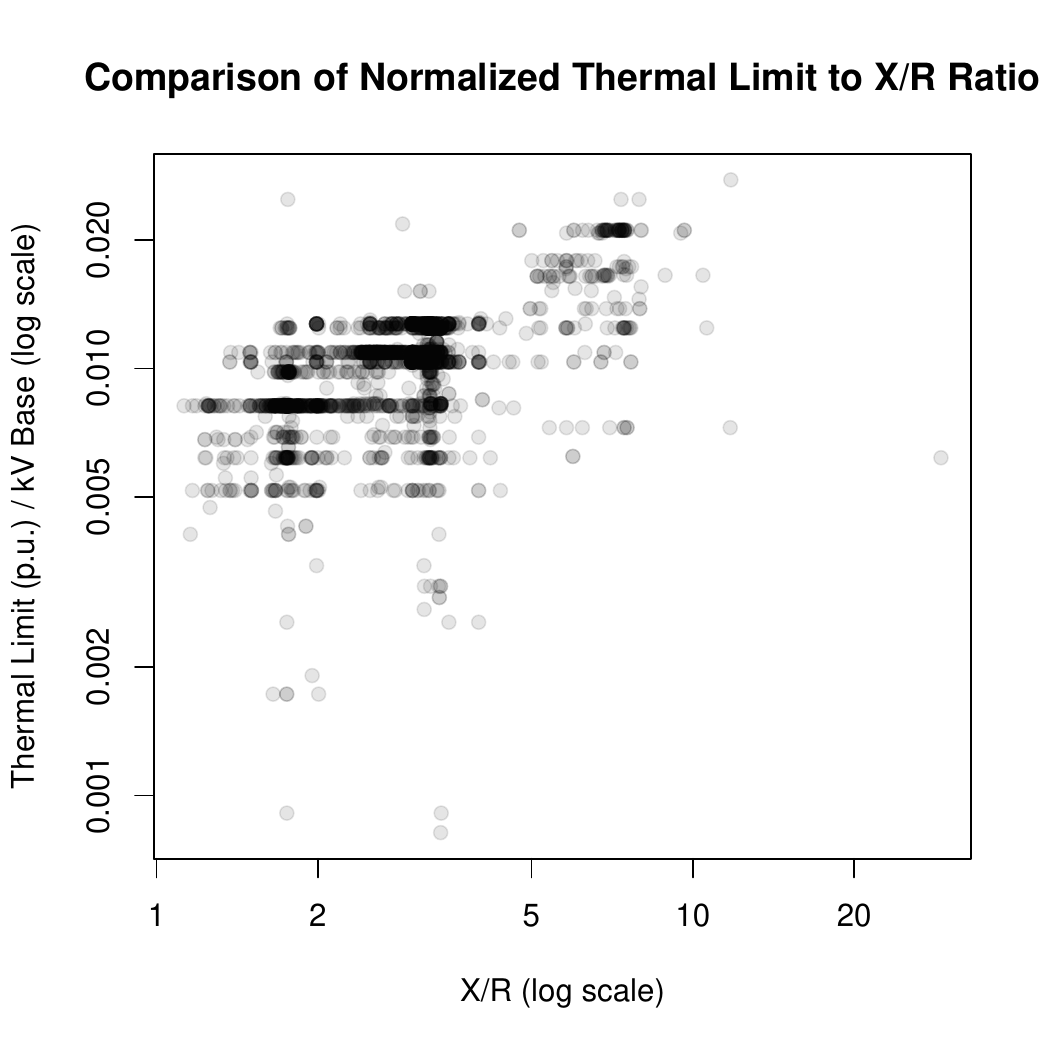}
    \includegraphics[width=6.5cm]{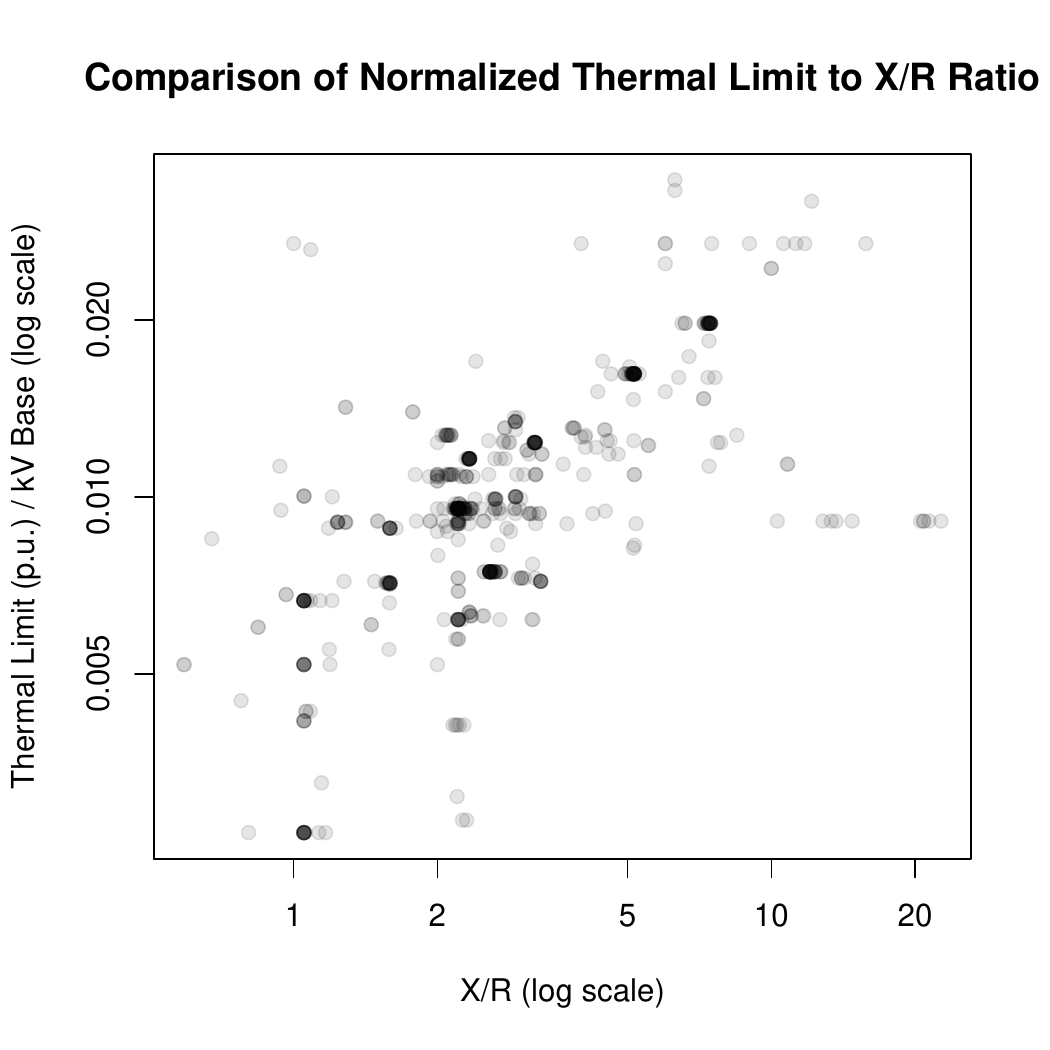} 
    \caption{Polish Network case2383wp (left) and EIRGrid Summer Night Valley 2013 (right).}
\label{fig:ncap_vs_xr}
\end{figure}

\begin{figure}[h!]
\center
    \includegraphics[width=6.5cm]{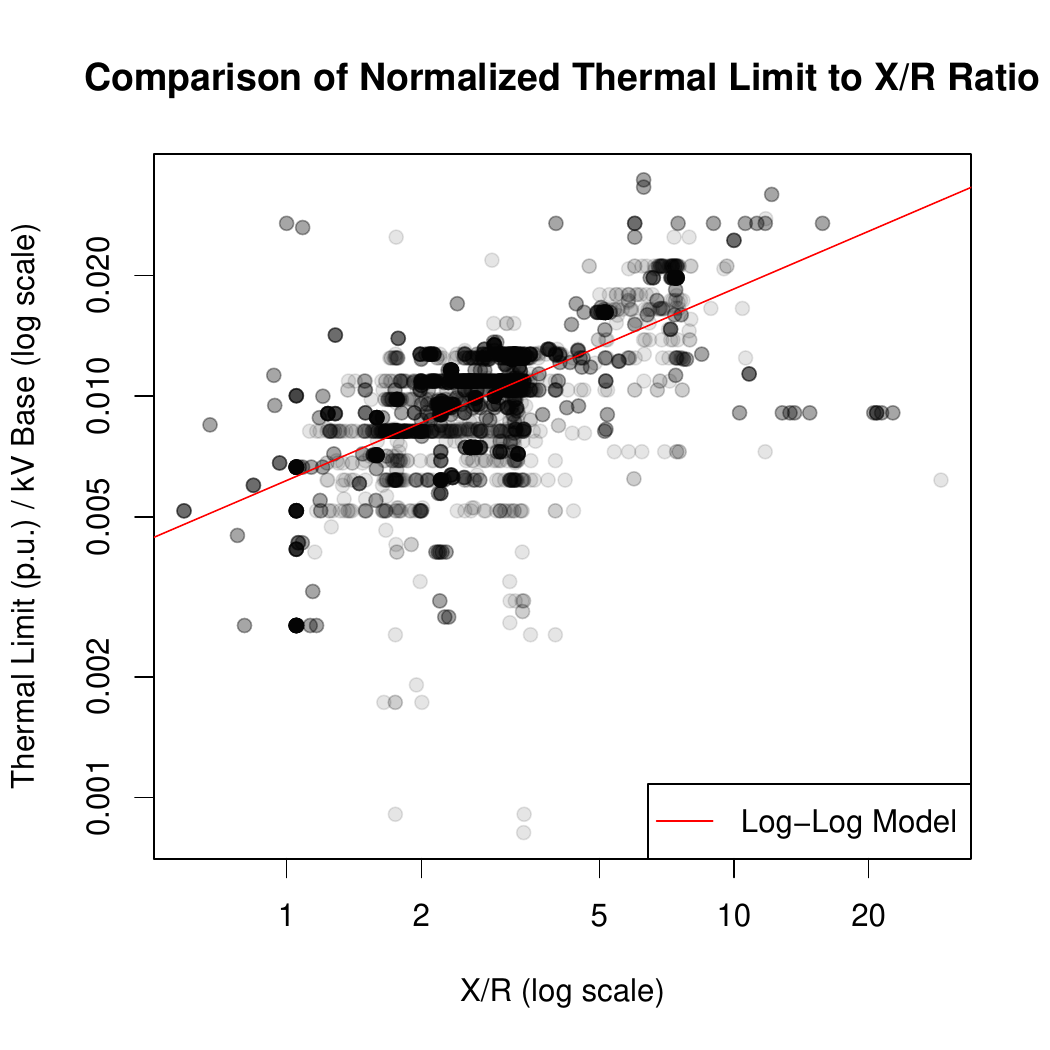}
    
    \begin{tabular}{|r|r|r|r|r||r|r|r|r|r|r|r|r|r|c|c|}
    \hline
    $a$ & $k$ \\
    \hline
    \hline
    -5.0886 & 0.4772 \\
    \hline
    \end{tabular}
    \label{tbl:line:stat}

    \caption{Linear Log-Log Model for equal proportions of data from the Polish Network case2383wp and EIRGrid Summer Night Valley 2013.}
\label{fig:ncap_vs_xr_fit}
\end{figure}

Our statistical model for line capacities was inspired by two key intuitions:
\begin{enumerate}
\item Thermal limits in AC power networks are typically given in terms of power flow (i.e. MVA), rather than current flow (i.e. MA).  This leads to a property that conductors of the same type and configuration can have different MVA limits depending on their nominal voltage, despite the physical properties of the line being identical.  If the MVA limits are converted to MA limits using the nominal bus voltage, equal conductors should have similar thermal limits.
\item The ratio of resistance to impedance may provide some insight into the line's conductor type and configuration, since this ratio should be independent of the line length, whereas the values taken individually will be proportional to the length.
\end{enumerate}
%
After plotting these two values on a log scale, we observe a positive correlation, as illustrated in Figure \ref{fig:ncap_vs_xr}.  We then combine these two data sets in equal proportions (note that the Polish network produced 2600 data points while the EIRGrid network only produced 650), and fit a linear regression model to the log-log data, which fits parameters $a$ and $k$ in the following function,
 \begin{align}
y &=   \bm e^{a} x^{k} \label{eq:log_log_model}
\end{align}
The fit log-log model and its parameters are presented in Figure \ref{fig:ncap_vs_xr_fit}.  We recognize that this linear fit is fairly crude, however, the results in Section \ref{sec:nesta} demonstrate that it is sufficient for generating optimization test cases.  It is also interesting to notice that the model selects $k=0.4772$, which is nearly a square root.  This is an encouraging result as thermal limits are closely related to heat dissipation and line losses, which increase quadratically with the line's current (i.e. ${loss}^p_{ij} = \bm r_{ij} e_{ij}^2$).

\noindent
This model (TL-Stat) can be used to estimate thermal limits in the following way.  For any line with the necessary parameters,
$r$, $x$, $\dot{v}$ (on both sides of the line), the per unit thermal limit $t$ is estimated as,
\begin{align}
t = \dot{v} \bm e^{-5.0886} \left( \frac{x}{r} \right)^{0.4772}
\end{align}

\subsubsection{A Reasonable Upper Bound}

Although the statistical model for line thermal limits is quite useful, it cannot be applied in cases where data for $r$, $x$, or $\dot{v}$ is missing.  Notable examples include: transformers, where the nominal voltage value differs on both sides of the line, and ideal lines, which do not have an $r$ value.  For these cases, it is helpful to have a method for producing reasonable upper bounds on the thermal limit.  Typically, when no thermal limit is known, a large value is used in its place, such as 9.9 p.u.\ in the \matpower case and 100.0 p.u.\ in the EIRGrid case.  The key problem with this approach is that it most often deactivates the line thermal limit constraint and hides the true maximum throughput of the transmission line.  Here we develop a theoretical approach for setting a line's thermal limit to a value that is close to the theoretical maximum but does not deactivate the thermal limit constraint, and which accurately indicates the line's true throughput limitations.

Recall that the thermal limit of a line is specified on the apparent power of the line, that is,
\begin{align}
p_{ij}^2 + q_{ij}^2 \leq \boldsymbol t_{ij}^2
\end{align}
Our goal is to find a reasonable upper bound for $\boldsymbol t_{ij}^2$.  Observe that the value of $t_{ij}^2$ can be stated in terms of the the voltage variables as follows,
\begin{align}
t_{ij}^2 &= v_i^2 e_{ij}^2 \\
t_{ij}^2 &= v_i^2 \boldsymbol y_{ij}^2 |V_i - V_j|^2 \\
t_{ij}^2 &= v_i^2 \boldsymbol y_{ij}^2 (v_i^2 + v_j^2 - 2v_iv_j\cos(\theta_i - \theta_j) ) \label{eq:thermal:limit:obj}
\end{align}
If we have reasonable bounds on $v_i, v_j, \theta_i - \theta_j$, then the task of determining the maximum throughput of a line can be stated as a numerical maximization problem where the objective is \eqref{eq:thermal:limit:obj} and the constraints are the voltage bounds.  However, given bounds on the voltage magnitudes ($\boldsymbol {v^l}, \boldsymbol {v^u}$) and the line's phase angle difference ($\boldsymbol \theta_{ij}^\Delta$) we observe that the optimal point of this maximization problem is often,
 \begin{align}
\boldsymbol t_{ij}^2 &= ({\boldsymbol {v^u}_i})^2 \boldsymbol y_{ij}^2  (({\boldsymbol {v^u}_i})^2 + ({\boldsymbol {v^u}_j})^2 - 2\boldsymbol {v^u}_i\boldsymbol {v^u}_j\cos(\boldsymbol \theta_{ij}^\Delta) ) \label{eq:thermal:limit}
\end{align}
in realistic lines.\footnote{The key assumption being that $2 \boldsymbol y \sin(\boldsymbol \theta^\Delta / 2)$ is large compared to $ \boldsymbol {v^u} - \boldsymbol {v^l}$.}  All AC networks have reasonable bounds on $\boldsymbol {v^l}, \boldsymbol {v^u}$ and recent studies \cite{PES2,Purchala:2005gt} have suggested $15^\circ$ is a reasonable value for $\boldsymbol \theta_{ij}^\Delta$.  Combining these values with \eqref{eq:thermal:limit} we have a model for a reasonable upper bound on a line's thermal limit (TL-UB).

\subsubsection{Validating the Models}

\begin{figure}[h!]
\center
    \includegraphics[width=6.5cm]{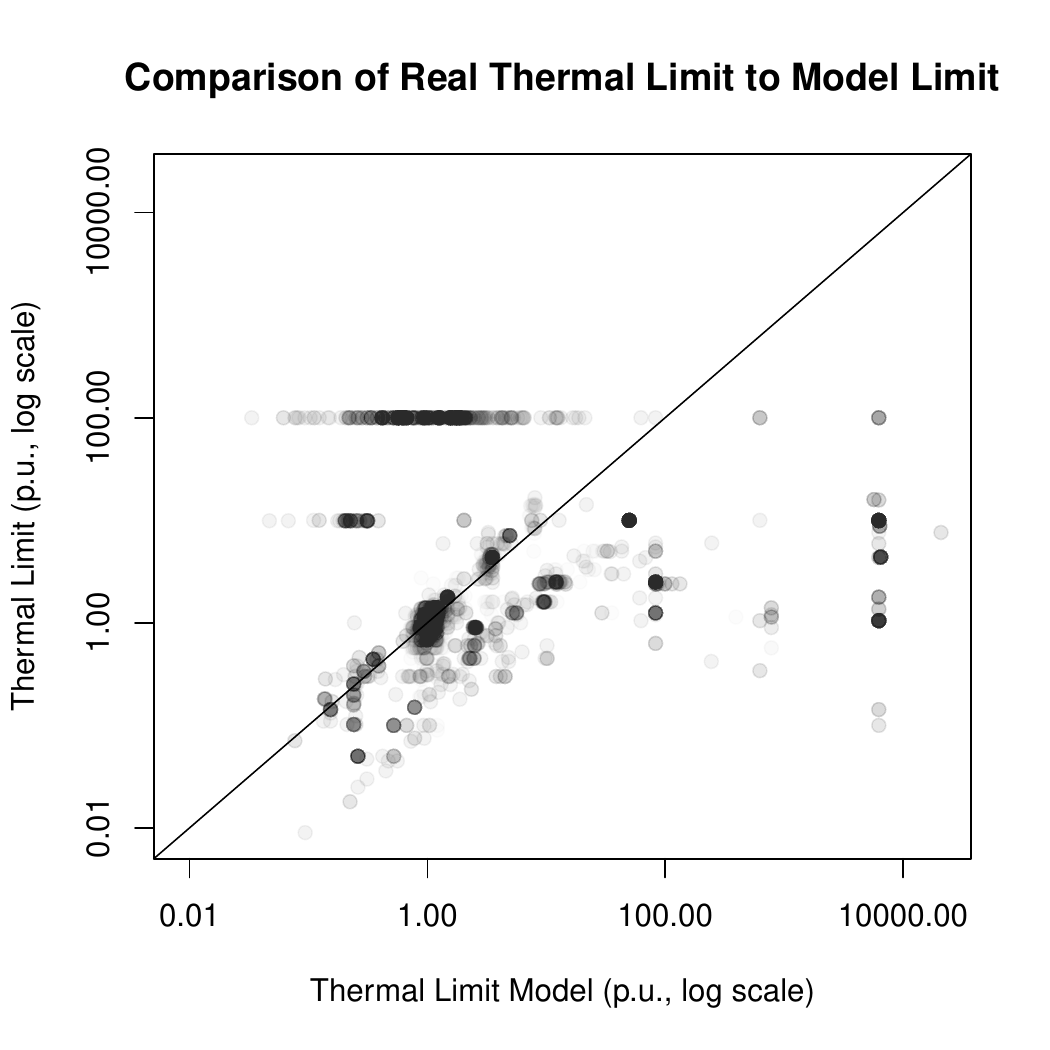}
    \includegraphics[width=6.5cm]{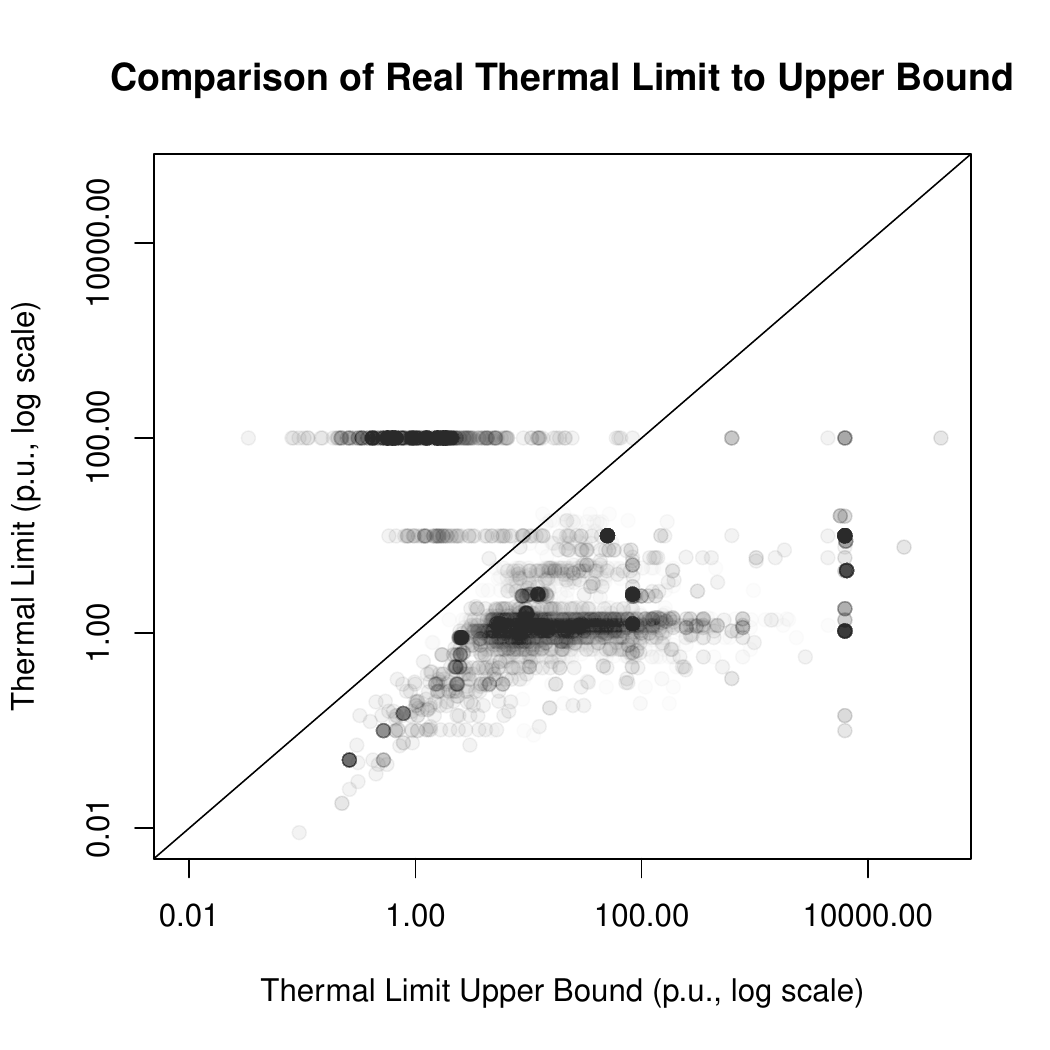}
    \caption{Comparison of the Statical Model (left) and the Thermal Limit Upper Bound Calculation (right) to the Given MVA Limits for Equal Proportions of the Polish Network case2383wp and EIRGrid Summer Night Valley 2013 (no data filtering used).}
\label{fig:cap_vs_ub}
\end{figure}

To validate that both the statical and upper bound models are producing reasonable values, Figure \ref{fig:cap_vs_ub} plots the correlation of the proposed MVA values to the real line thermal limits given in the Polish and EIRGrid networks.  Both plots exhibit horizontal and vertical lines of points near the boundary of the figure, which are notable special cases.  The collections of points forming horizontal lines around $y=9.9$ p.u. and $y=100$ p.u., are indicating missing thermal limits in the Polish and EIRGrid data respectively.  The vertically aligned points around $x=10000$ are a collection of lines with a very small reactance value modeling an ideal conductor, hence both models produce a value similar to positive infinity for their thermal limit.  A perfect predictive model would place all of the points on the line $x=y$.  The small clouds of points on that line in the left figure, indicate that the statistical model is producing reasonable values for many lines.  In contrast, the upper bound model is consistently below the $x=y$ line, indicating it is in fact an optimistically high bound on the line's flow capacity, as it was designed to be.

\subsubsection{A Complete Thermal Limit Model}

A robust thermal limit model (TL-Stat) is developed by combining the statistical model and the upper bound model in the following way.  
A line's thermal limit is computed with both the statistical model and thermal limit upper bound model (TL-UB) and the minimum of these two values
is used.  If the data required for the statistical model is unavailable on a given line, only thermal limit upper bound is computed.  For test cases that include some line thermal limits, the model is applied only when it produces a tighter limit than the specified value.



\section{The NESTA Networks}
\label{sec:nesta}

Now that we have built statical models for completing the missing data in the traditional test cases, we are ready to build the NESTA networks.  In designing the NESTA networks we strive to keep the data as realistic and accurate as possible.  Hence we only resort to using the statistical models when no realistic data is provided.  Table \ref{tbl:nesta:inst} summarizes which models are used to convert existing test cases into NESTA test cases.

Recent research \cite{PES2, LPAC_ijoc, QCarchive} has indicated that phase angle difference bounds may be valuable constraints for power system optimization.  Hence all of the NESTA models have a generous phase angle difference bound of $\bm {\theta^\Delta}_{ij} = 30^\circ$.

The statistical models were not applied to the PEGASE Test Cases for two reasons.  First, due to the aggregation of network components, these  statical models are less meaningful in these networks.  And second, given the recent studies of these networks in \cite{6488772,rte_pegase}, there is significant value in keeping them relatively unchanged.  It was observed that the phase angle bound and thermal limit models did not change the solution of these networks produced by \matpower.

Most of the time the NESTA test cases lead to feasible AC-OPF solutions, however, in a few cases it is necessary to override the statistical models to produce AC-OPF feasibility. The special overrides column in Table \ref{tbl:nesta:inst} indicates which networks require additional modifications.  These modifications are detailed as follows:

\begin{table}[h!]
\center
\footnotesize
\begin{tabular}{|r||r||c|c|c|c|c|c|r|r|r|r|r|r|r|c|c|}
\hline
 &    & \multicolumn{4}{c|}{Model} &  &  \\
NESTA & Original & Active &  Reactive   & Gen.  & Thermal  &  Phase Angle & Special \\
Name &   Name & Gen. & Gen.  & Cost  & Limit &  Bound & Overrides \\
\hline
\hline
\multicolumn{8}{|c|}{IEEE Power Flow Test Cases}\\
\hline
nesta\_case14\_ieee &14 Bus & AG-Stat & RG-AM50 & AC-Stat & TL-Stat & $30^\circ$ & --- \\
\hline
nesta\_case30\_ieee &30 Bus & AG-Stat & RG-AM50 & AC-Stat & TL-Stat & $30^\circ$ & ---\\
\hline
nesta\_case57\_ieee & 57 Bus & AG-Stat & RG-AM50 & AC-Stat & TL-Stat & $30^\circ$ & ---\\
\hline
nesta\_case118\_ieee & 118 Bus & AG-Stat & RG-AM50 & AC-Stat & TL-Stat & $30^\circ$ & ---\\
\hline
nesta\_case300\_ieee & 300 Bus & AG-Stat & RG-AM50 & AC-Stat & TL-UB & $30^\circ$ & ---\\
\hline
\hline
\multicolumn{8}{|c|}{IEEE Dynamic Test Cases}\\
\hline
nesta\_case162\_ieee\_dtc & 17 Generator & AG-Stat & RG-AM50 & AC-Stat & TL-Stat & $30^\circ$ & --- \\
\hline
\hline
\multicolumn{8}{|c|}{IEEE Reliability Test Systems (RTS)}\\
\hline
nesta\_case24\_ieee\_rts &  RTS-79 & --- & --- & --- & --- & $30^\circ$ & --- \\
\hline
nesta\_case73\_ieee\_rts & RTS-96 &--- & --- & --- & --- & $30^\circ$ & --- \\
\hline
\hline
\multicolumn{8}{|c|}{Matpower Test Cases}\\
\hline
nesta\_case2383wp\_mp & case2383wp & --- & --- & --- & --- & $30^\circ$ & --- \\
\hline
nesta\_case2736sp\_mp & case2736sp & --- & --- & --- & --- & $30^\circ$ & ---\\
\hline
nesta\_case2737sop\_mp & case2737sop & --- & --- & --- & --- & $30^\circ$ & --- \\
\hline
nesta\_case2746wop\_mp & case2746wop & --- & --- & --- & --- & $30^\circ$ & --- \\
\hline
nesta\_case2746wp\_mp & case2746wp & --- & --- & --- & --- & $30^\circ$ & --- \\
\hline
nesta\_case3012wp\_mp & case3012wp & --- & --- & --- & TL-Stat & $30^\circ$ & --- \\
\hline
nesta\_case3120sp\_mp & case3120sp & --- & --- & --- & TL-Stat & $30^\circ$ & --- \\
\hline
nesta\_case3375wp\_mp & case3375wp & --- & RG-AM50 & --- & TL-Stat & $30^\circ$ & Yes \\
\hline
\hline
\multicolumn{8}{|c|}{PEGASE Test Cases}\\
\hline
nesta\_case89\_pegase & case89pegase & --- & --- & --- & TL-UB & $30^\circ$ & --- \\
\hline
nesta\_case1354\_pegase & case1354pegase & --- & --- & --- & TL-UB & $30^\circ$ & --- \\
\hline
nesta\_case2869\_pegase & case2869pegase & --- & --- & --- & TL-UB & $30^\circ$ & --- \\
\hline
nesta\_case9241\_pegase & case9241pegase & --- & --- & --- & TL-UB & $30^\circ$ & --- \\
\hline
\hline
\multicolumn{8}{|c|}{Edinburgh Test Case Archive}\\
\hline
nesta\_case186\_edin & Iceland & AG-Stat & RG-AM50 & AC-Stat & TL-Stat & $30^\circ$ & --- \\
\hline
nesta\_case2224\_edin & Great Britain & --- & --- & AC-Stat & TL-Stat & $30^\circ$ & Yes \\
\hline
nesta\_case29\_edin & Reduced G.B. &  --- & --- & AC-Stat & TL-UB & $30^\circ$ & --- \\
\hline
%
\hline
\multicolumn{8}{|c|}{EIRGrid Test Cases}\\
\hline
nesta\_case1394sop\_eir & Summer Valley 2013 & --- & --- & AC-Stat & TL-UB & $30^\circ$ & ---  \\
\hline
nesta\_case1397sp\_eir & Summer Peak 2013 & --- & --- & AC-Stat & TL-UB & $30^\circ$ & --- \\
\hline
nesta\_case1460wp\_eir & Winter Peak 2013-14 & --- & --- & AC-Stat & TL-UB & $30^\circ$ & ---  \\
\hline
\hline
\multicolumn{8}{|c|}{Publication Test Cases}\\
\hline
nesta\_case3\_lmbd & 3-Bus & --- & --- & --- & --- & $30^\circ$ & --- \\
\hline
nesta\_case4\_gs & case4gs & ---  & --- & AC-Stat & TL-Stat & $30^\circ$ & --- \\
\hline
nesta\_case5\_pjm & case5 & ---  & --- & --- & TL-Stat & $30^\circ$ & --- \\
\hline
nesta\_case6\_ww & case6ww & --- & --- & --- & --- & $30^\circ$ & --- \\
\hline
nesta\_case6\_c & case6bus & AG-Stat & RG-AM50 & AC-Stat & TL-Stat & $30^\circ$ & --- \\
\hline
nesta\_case9\_wscc & case9 &  --- & --- & --- & --- & $30^\circ$ & --- \\
\hline
nesta\_case30\_as & 30 Bus-as & --- & --- & --- & --- & $30^\circ$ & --- \\
\hline
nesta\_case30\_fsr & 30 Bus-fsr & --- & --- & --- & --- & $30^\circ$ & --- \\
\hline
nesta\_case39\_epri & case39 & --- & --- & --- & --- & $30^\circ$ & ---  \\
\hline
\end{tabular}
\caption{NESTA Instance Generation Details}
\label{tbl:nesta:inst}
\end{table}

\paragraph{IEEE 300 Bus Test Case}
As mentioned previously, this network exhibits atypical large voltage drops.  As a result, the statistical line flow model (TL-Stat) is too constraining for this network.  To ensure the network's feasibility, we resort to using the line flow upper bounds model (TL-UB) for the thermal limit on all of the lines.

\paragraph{Reduced Great Britain Test Case}
Many of the lines in this network have thermal limits well above the line flow upper bounds model.  The model is used to assign reasonable values to these lines. 

\paragraph{Great Britain Test Case}
At its default value, the thermal limit constraints on branches 599--592 and 2162--1868 make this model infeasible.  Simply increasing these thermal limits to 800 and 50 MVA respectively resolves the issue and makes the network feasible.

\paragraph{Polish 3375 Winter Peak Test Case}
Unlike the other Polish network cases, this test case exhibits a number of strange features, including generators with unbounded reactive power capabilities and transmission lines without thermal limits.  After applying the specified models, the following modifications are needed to make the test case feasible:  The reactive capability of the first generator is increased to 4000 MVar to support a similar sized reactive load at bus 10079.   Lines 10082-10079, and 10129-10130 have their thermal limits increased to 1000 MVA and lines 10330-10359 and 10331-10359 have their thermal limits increased to 250 MVA.  Given the scale of this network, these modifications are quite minor.

\subsection{Results}
\label{sec:results}

To verify the usefulness of NESTA for optimization applications we revisit the power flow relaxations from Section \ref{sec:motivation} in Table \ref{tbl:ac:nesta:bounds}.  In this table ``err.'' indicates the solver failed to converge while ``---'' indicates the model was not applicable to that input data.
Note that CP relaxation cannot be applied to networks with negative $r,x$ values and that the SDP-OPF implementation does not support inactive buses (i.e. type 4 in  \matpower), which occur in the EIRGrid test cases.

In Table \ref{tbl:ac:nesta:bounds}, many of the best optimality gaps have remained small (i.e. below 1\%).  This is to be expected, as a number of the test cases were not modified significantly from their original form.  However, if we focus our attention on the NF+LL results, a number of networks do exhibit significant optimality gaps, such as nesta\_case30\_ieee,  nesta\_case162\_ieee\_dtc, nesta\_case300\_ieee, and nesta\_case2224\_edin.  This suggests that some benefits may be gained from solving the AC-OTS problem on these networks.  

\begin{table}[h!]
\center
\begin{tabular}{|r||r||r|r|r|r|r|r|r|r|r||r|r|r|r|c|c|}
\hline
                   & \$/h & \multicolumn{4}{c|}{Optimality Gap (\%)}  \\
Test Case & AC  & CP & NF+LL & SOC & SDP \\
\hline
\hline
 nesta\_case3\_lmbd & 5812.64 & 2.99 & 2.26 & 1.32 & 0.39 \\
\hline
 nesta\_case4\_gs & 156.43 & 29.69 & 7.01 & 0.00 & \bf 0.00  \\
\hline
 nesta\_case5\_pjm & 17551.89 & 15.62 & 14.55 & 14.54 & 5.22 \\
\hline
 nesta\_case6\_c & 23.21 & 1.88 & 0.36 & 0.30 & 0.00 \\
\hline
 nesta\_case6\_ww & 3143.97 & 3.10 & 0.81 & 0.63 & \bf 0.00  \\
\hline
 nesta\_case9\_wscc & 5296.69 & 1.52 & 0.17 & 0.00 & 0.00 \\
\hline
 nesta\_case14\_ieee & 244.05 & 5.18 & 0.28 & 0.11 & \bf 0.00  \\
\hline
 nesta\_case24\_ieee\_rts & 63352.20 & 3.71 & 0.36 & 0.01 & \bf 0.00  \\
\hline
 nesta\_case29\_edin & 29895.49 & 0.76 & 0.16 & 0.14 & \bf 0.00  \\
\hline
 nesta\_case30\_as & 803.13 & 4.42 & 0.38 & 0.06 & 0.00 \\
\hline
 nesta\_case30\_fsr & 575.77 & 1.83 & 0.50 & 0.39 & 0.00 \\
\hline
 nesta\_case30\_ieee & 204.97 & 27.91 & 16.69 & 15.88 & \bf 0.00  \\
\hline
 nesta\_case39\_epri & 96505.52 & 0.96 & 0.08 & 0.05 & 0.01 \\
\hline
 nesta\_case57\_ieee & 1143.27 & 1.59 & 0.27 & 0.06 & \bf 0.00  \\
\hline
 nesta\_case73\_ieee\_rts & 189764.08 & 3.56 & 0.36 & 0.03 & \bf 0.00  \\
\hline
 nesta\_case89\_pegase & 5819.81 & 1.50 & 0.36 & 0.17 & 0.00 \\
\hline
 nesta\_case118\_ieee & 3718.64 & 7.87 & 2.43 & 2.07 & 0.06 \\
\hline
 nesta\_case162\_ieee\_dtc & 4230.23 & 15.44 & 4.43 & 4.03 & 1.08 \\
\hline
 nesta\_case189\_edin & 849.29 & --- & 1.79 & 0.21 & 0.07 \\
\hline
 nesta\_case300\_ieee & 16891.28 & --- & 3.65 & 1.18 & 0.08 \\
\hline
 nesta\_case1354\_pegase & 74069.35 & 1.36 & 0.20 & 0.08 & 0.00$^\star$ \\
\hline
 nesta\_case1394sop\_eir & 1366.81 & 7.31 & 1.25 & 0.82 & ---  \\
\hline
 nesta\_case1397sp\_eir & 3888.99 & 5.55 & err. & 0.93 & ---  \\
\hline
 nesta\_case1460wp\_eir & 4640.18 & 41.26 & 1.25 & 0.89 & ---  \\
\hline
 nesta\_case2224\_edin & 38127.69 & 8.45 & 6.33 & 6.09 & 1.22 \\
\hline
 nesta\_case2383wp\_mp & 1868511.78 & 5.35 & 1.38 & 1.05 & 0.37 \\
\hline
 nesta\_case2736sp\_mp & 1307883.11 & 2.44 & 0.46 & 0.30 & 0.00$^\star$ \\
\hline
 nesta\_case2737sop\_mp & 777629.29 & 1.75 & 0.39 & 0.25 & 0.00$^\star$ \\
\hline
 nesta\_case2746wp\_mp & 1631775.07 & 3.09 & 0.53 & 0.32 & 0.00$^\star$ \\
\hline
 nesta\_case2746wop\_mp & 1208279.78 & 2.50 & 0.56 & 0.37 & 0.00$^\star$ \\
\hline
 nesta\_case2869\_pegase & 133999.29 & 1.16 & 0.19 & 0.09 & 0.00$^\star$ \\
\hline
 nesta\_case3012wp\_mp & 2600842.72 & --- & 5.18 & 1.02 & err. \\
\hline
 nesta\_case3120sp\_mp & 2145739.40 & --- & 5.26 & 0.55 & err. \\
\hline
 nesta\_case3375wp\_mp & 7435697.48 & --- & 2.01 & 0.52 & err. \\
\hline
 nesta\_case9241\_pegase & 315913.26 & --- & 2.16 & --- & --- \\
 \hline
\end{tabular}
\caption{AC-OPF Bounds on the NESTA Test Cases ($\star$ - solver reported numerical accuracy warnings).}
\label{tbl:ac:nesta:bounds}
\end{table}

\subsection{Building More Challenging Test Cases}
\label{sec:congested}

The first goal of NESTA is to establish a comprehensive archive of transmission system test cases with reasonable network parameter values.  However, its secondary goal is to serve as a basis for building modified networks for testing a variety of power system scenarios and applications.  In this section we demonstrate two of many possible ways that NESTA and the models developed in this report can be used to build interesting new test cases.

It was observed in \cite{pscc_ots, 6345676} that power flow congestion is a key component in interesting AC-OTS test cases.  This observation inspires the following approach to building congested NESTA test cases.  For each of the standard NESTA cases, we make the generator capabilities unbounded and then solve an {\em active power increase} optimization problem, which increases the active power demands proportionally throughout the network until the line thermal limits are binding.  Once a maximal increase in active power demand is determined, the set-points of the generators and loads specify an AC power flow test case.  The statistical models, AG-Stat, RG-AL50, and AC-Stat are then applied to these new power flow test cases to produce the Active Power Increase (API) test cases.  The optimization gaps of these test cases is detailed in Table \ref{tbl:ac:nesta:api:bounds}.  Interestingly, the congestion introduced by increasing the active power demands leads to significant optimality gaps. 


\begin{table}[h!]
\center
\begin{tabular}{|r||r||r|r|r|r|r|r|r|r|r||r|r|r|r|c|c|}
\hline
                   & \$/h & \multicolumn{4}{c|}{Optimality Gap (\%)}  \\
Test Case & AC  & CP & NF+LL & SOC & SDP \\
\hline
\hline
 nesta\_case3\_lmbd\_\_api & 367.74 & 14.79 & 8.16 & 3.30 & 1.26 \\
\hline
 nesta\_case4\_gs\_\_api & 767.27 & 6.72 & 0.77 & 0.65 & \bf 0.00  \\
\hline
 nesta\_case5\_pjm\_\_api & 2998.54 & 0.75 & 0.45 & 0.45 & \bf 0.00  \\
\hline
 nesta\_case6\_c\_\_api & 814.40 & 2.74 & 0.41 & 0.35 & \bf 0.00  \\
\hline
 nesta\_case6\_ww\_\_api & 273.76 & 17.17 & 13.64 & 13.33 & 0.00$^\star$ \\
\hline
 nesta\_case9\_wscc\_\_api & 656.60 & 16.01 & 0.52 & 0.00 & \bf 0.00  \\
\hline
 nesta\_case14\_ieee\_\_api & 325.56 & 8.89 & 1.78 & 1.34 & \bf 0.00  \\
\hline
 nesta\_case24\_ieee\_rts\_\_api & 6421.37 & 24.12 & 21.14 & 20.70 & 1.45 \\
\hline
 nesta\_case29\_edin\_\_api & 295782.68 & 0.75 & 0.44 & 0.44 & err. \\
\hline
 nesta\_case30\_as\_\_api & 571.13 & 8.01 & 5.14 & 4.76 & 0.00 \\
\hline
 nesta\_case30\_fsr\_\_api & 372.14 & 48.80 & 46.19 & 45.97 & 11.06 \\
\hline
 nesta\_case30\_ieee\_\_api & 415.53 & 12.75 & 1.48 & 1.01 & 0.00 \\
\hline
 nesta\_case39\_epri\_\_api & 7466.25 & 13.31 & 3.71 & 2.99 & \bf 0.00  \\
\hline
 nesta\_case57\_ieee\_\_api & 1430.65 & 3.51 & 0.46 & 0.21 & 0.08 \\
\hline
 nesta\_case73\_ieee\_rts\_\_api & 20123.98 & 17.83 & 15.63 & 14.34 & 4.29 \\
\hline
 nesta\_case89\_pegase\_\_api & 4288.02 & 22.60 & 20.71 & 20.43 & 18.11 \\
\hline
 nesta\_case118\_ieee\_\_api & 10325.27 & 49.69 & 44.35 & 44.08 & 31.50 \\
\hline
 nesta\_case162\_ieee\_dtc\_\_api & 6111.68 & 19.39 & 1.97 & 1.34 & 0.85 \\
\hline
 nesta\_case189\_edin\_\_api & 1982.82 & --- & 8.21 & 5.78 & 0.05 \\
\hline
 nesta\_case300\_ieee\_\_api & 22866.01 & --- & 1.81 & 0.84 & 0.00 \\
\hline
 nesta\_case1354\_pegase\_\_api & 59920.94 & 3.33 & 0.74 & 0.56 & 0.20$^\star$ \\
\hline
 nesta\_case1394sop\_eir\_\_api & 3176.26 & 15.07 & err. & 1.58 & --- \\
\hline
 nesta\_case1397sp\_eir\_\_api & 5983.48 & 10.13 & err. & 1.59 & ---  \\
\hline
 nesta\_case1460wp\_eir\_\_api & 6262.09 & 9.32 & 1.26 & 0.87 & ---  \\
\hline
 nesta\_case2224\_edin\_\_api & 46235.43 & 9.07 & 2.90 & 2.77 & 1.10 \\
\hline
 nesta\_case2383wp\_mp\_\_api & 23499.48 & 3.10 & 1.24 & 1.12 & 0.10 \\
\hline
 nesta\_case2736sp\_mp\_\_api & 25437.70 & 3.89 & 1.45 & 1.33 & 0.07 \\
\hline
 nesta\_case2737sop\_mp\_\_api & 21192.40 & 4.62 & 1.20 & 1.06 & 0.00 \\
\hline
 nesta\_case2746wp\_mp\_\_api & 27291.58 & 2.31 & 0.68 & 0.58 & 0.00 \\
\hline
 nesta\_case2746wop\_mp\_\_api & 22814.86 & 1.87 & 0.59 & 0.49 & 0.00 \\
\hline
 nesta\_case2869\_pegase\_\_api & 96573.10 & 5.16 & 1.68 & 1.49 & 0.92$^\star$ \\
\hline
 nesta\_case3012wp\_mp\_\_api & 27917.36 & --- & 3.51 & 0.90 & err. \\
\hline
 nesta\_case3120sp\_mp\_\_api & 22874.98 & --- & 6.26 & 3.03 & err. \\
\hline
 nesta\_case3375wp\_mp\_\_api & 48898.95 & --- & 2.08 & err. & 0.00$^\star$ \\
\hline
 nesta\_case9241\_pegase\_\_api & 241975.18 & --- & 3.12 & 2.59 & --- \\
\hline
\end{tabular}
\caption{AC-OPF Bounds on the NESTA-API Test Cases ($\star$ - solver reported numerical accuracy warnings).}
\label{tbl:ac:nesta:api:bounds}
\end{table}

An entirely different approach to modifying the NESTA cases is inspired by the recent lines of research \cite{PES2, LPAC_ijoc, QCarchive} that indicate phase angle difference bounds can have significant impacts on power system optimization approaches.  To investigate these impacts, new test cases are constructed in the following way.  For each of the standard NESTA cases, a {\em small angle difference} optimization problem is solved, which finds the minimum value of $\bm {\theta^\Delta}$ that can be applied on all of the lines in the network, while retaining a feasible AC power flow.  Once the small, but feasible, value of $\bm {\theta^\Delta}$ is determined, the original test case is updated with this value, yielding new optimal power flow test case, which we call the Small Angle Difference (SAD) test cases.  The optimality gaps of these test cases are detailed in Table \ref{tbl:ac:nesta:sad:bounds}.  Interestingly, this entirely different approach to making challenging test cases also leads to significant optimality gaps.

%

\begin{table}[h!]
\center
\begin{tabular}{|r||r||r|r|r|r|r|r|r|r|r||r|r|r|r|c|c|}
\hline
                   & \$/h & \multicolumn{4}{c|}{Optimality Gap (\%)}  \\
Test Case & AC  & CP & NF+LL & SOC & SDP \\
\hline
\hline
 nesta\_case3\_lmbd\_\_sad & 5992.72 & 5.90 & 5.15 & 4.28 & 2.06 \\
\hline
 nesta\_case4\_gs\_\_sad & 324.02 & 66.06 & 4.93 & 4.90 & 0.05 \\
\hline
 nesta\_case5\_pjm\_\_sad & 26423.32 & 43.95 & 5.37 & 3.61 & \bf 0.00  \\
\hline
 nesta\_case6\_c\_\_sad & 24.43 & 6.79 & 1.43 & 1.36 & 0.00 \\
\hline
 nesta\_case6\_ww\_\_sad & 3149.51 & 3.27 & 0.98 & 0.80 & \bf 0.00  \\
\hline
 nesta\_case9\_wscc\_\_sad & 5590.09 & 6.69 & 4.20 & 1.50 & 0.00 \\
\hline
 nesta\_case14\_ieee\_\_sad & 244.15 & 5.22 & 0.26 & 0.06 & 0.00 \\
\hline
 nesta\_case24\_ieee\_rts\_\_sad & 79804.96 & 23.56 & 19.44 & 11.42 & 6.05 \\
\hline
 nesta\_case29\_edin\_\_sad & 46933.26 & 36.79 & 34.98 & 34.47 & 28.44 \\
\hline
 nesta\_case30\_as\_\_sad & 914.44 & 16.06 & 10.78 & 9.16 & 0.47 \\
\hline
 nesta\_case30\_fsr\_\_sad & 577.73 & 2.17 & 0.77 & 0.62 & 0.07 \\
\hline
 nesta\_case30\_ieee\_\_sad & 205.11 & 27.96 & 6.48 & 5.84 & \bf 0.00  \\
\hline
 nesta\_case39\_epri\_\_sad & 97219.04 & 1.69 & 0.57 & 0.11 & 0.09 \\
\hline
 nesta\_case57\_ieee\_\_sad & 1143.88 & 1.64 & 0.32 & 0.11 & 0.02 \\
\hline
 nesta\_case73\_ieee\_rts\_\_sad & 235241.70 & 22.21 & 16.49 & 8.37 & 4.10 \\
\hline
 nesta\_case89\_pegase\_\_sad & 5827.01 & 1.62 & 0.48 & 0.28 & 0.03 \\
\hline
 nesta\_case118\_ieee\_\_sad & 4324.17 & 20.77 & 14.34 & 12.89 & 7.57 \\
\hline
 nesta\_case162\_ieee\_dtc\_\_sad & 4369.19 & 18.13 & 7.47 & 7.08 & 3.65 \\
\hline
 nesta\_case189\_edin\_\_sad & 914.61 & --- & 7.92 & 2.25 & 1.20$^\star$ \\
\hline
 nesta\_case300\_ieee\_\_sad & 16910.23 & --- & 3.75 & 1.26 & 0.13 \\
\hline
 nesta\_case1354\_pegase\_\_sad & 74072.33 & 1.37 & 0.20 & 0.08 & 0.00$^\star$ \\
\hline
 nesta\_case1394sop\_eir\_\_sad & 1577.59 & 19.69 & 12.83 & 11.93 & ---  \\
\hline
 nesta\_case1397sp\_eir\_\_sad & 4582.08 & 19.84 & --- & 14.04 & ---  \\
\hline
 nesta\_case1460wp\_eir\_\_sad & 5367.75 & 49.22 & err. & 1.08 & ---  \\
\hline
 nesta\_case2224\_edin\_\_sad & 38385.14 & 9.06 & 6.61 & 6.18 & 1.22 \\
\hline
 nesta\_case2383wp\_mp\_\_sad & 1935308.12 & 8.62 & 4.48 & 4.00 & 1.30 \\
\hline
 nesta\_case2736sp\_mp\_\_sad & 1337042.77 & 4.56 & 2.51 & 2.34 & 2.18$^\star$ \\
\hline
 nesta\_case2737sop\_mp\_\_sad & 795429.36 & 3.95 & 2.57 & 2.42 & 2.24$^\star$ \\
\hline
 nesta\_case2746wp\_mp\_\_sad & 1672150.46 & 5.43 & 2.73 & 2.44 & 2.41$^\star$ \\
\hline
 nesta\_case2746wop\_mp\_\_sad & 1241955.30 & 5.14 & 3.14 & 2.94 & 2.71$^\star$ \\
\hline
 nesta\_case2869\_pegase\_\_sad & 134087.47 & 1.22 & 0.25 & 0.15 & 0.04$^\star$ \\
\hline
 nesta\_case3012wp\_mp\_\_sad & 2635451.29 & --- & 6.36 & 2.12 & err. \\
\hline
 nesta\_case3120sp\_mp\_\_sad & 2203807.23 & --- & 7.33 & 2.79 & err. \\
\hline
 nesta\_case3375wp\_mp\_\_sad & 7436381.61 & --- & 2.02 & 0.52 & --- \\
\hline
 nesta\_case9241\_pegase\_\_sad & 315932.06 & --- & 2.17 & 1.75 & --- \\
\hline
\end{tabular}
\caption{AC-OPF Bounds on the NESTA-SAD Test Cases ($\star$ - solver reported numerical accuracy warnings).}
\label{tbl:ac:nesta:sad:bounds}
\end{table}

In both Table \ref{tbl:ac:nesta:api:bounds} and Table \ref{tbl:ac:nesta:sad:bounds}, the significant optimality gaps may be caused by the non-convexities in the AC-OPF, which the relaxations do not capture, or by the AC heuristic finding local optimal solutions \cite{6581918}.  In either case, these modified test cases present interesting challenges for optimization algorithms.  

\section{Conclusions}
\label{sec:conclusion}

This report has highlighted the shortcomings of using existing test cases for studies of the AC Optimal Power Flow problem, and has proposed data driven models for improving these test cases.  The resulting test case archive, NESTA, has reasonable values for key power network parameters, including generation capabilities, costs, and line thermal limits.  Our validation demonstrates that many of the NESTA networks have significant optimality gaps, and are therefore interesting AC-OPF test cases.  Furthermore, two congested network case studies illustrated how NESTA and the proposed data driven models can be used as building blocks for constructing more challenging AC power flow optimization test cases.

It is important to note that while the research community's primary challenge has been to acquire the real-world data necessary to solve classic academic AC-OPF test cases, this is just the beginning of the data required for comprehensive studies of AC power flow optimization problems \cite{real_opf}.  Whenever possible the research community should try to acquire additional data such as: information about configurable assets including bus shunts, switches, and transformer tap settings; the N-1 contingency cases of interest to the network operator and designers; additional line thermal limits for short and long term overloading; ZIP load models; time series data for loads; and detailed generator models including historical market bids, ramp rates, startup/shutdown costs, and power capability curves.  Some of these details are available in the EIRGrid dataset \cite{EIRGrid} and the PES OPF Competition \cite{pes_opf_comp}, however they were not included in NESTA as the \matpower test case file format does not support these additional details.

Although NESTA has significant advantages over other readily available test cases, it is still far from a detailed real-world network specification.  This report has highlighted the need for deeper industry engagement for the development of more modern, detailed, and realistic test cases.  We hope that NESTA may serve as a stop gap for the research community to study various power system optimization problems, while more real-world test cases are being developed.

\vspace{-0.2cm}
\section{Acknowledgements}
NICTA is funded by the Australian Government through the Department of Communications and the Australian Research Council through the ICT Centre of Excellence Program. 

\bibliographystyle{plain}
\bibliography{nesta}

\end{document}